\documentclass{article}

\usepackage[preprint]{neurips_2026}

\usepackage[utf8]{inputenc} 
\usepackage[T1]{fontenc}    
\usepackage{url}            
\usepackage{booktabs}       
\usepackage{amsfonts}       
\usepackage{nicefrac}       
\usepackage{microtype}      
\usepackage{xcolor}         
\usepackage{microtype}
\usepackage{graphicx, xcolor}
\usepackage{subcaption}
\usepackage{booktabs} 
\usepackage{tcolorbox}
\usepackage{courier}
\usepackage{mathrsfs}
\usepackage{algorithm}
\usepackage{algpseudocode}
\usepackage{titletoc}

\graphicspath{{figure/}}

\newcommand{\nc}{\newcommand}
\nc{\dc}{\definecolor}
\dc{gsublue}{RGB}{0, 57, 166}
\dc{watpink}{RGB}{198, 0, 120}
\dc{watyellow}{RGB}{254, 211, 76}
\dc{watyellow2}{RGB}{235,171, 0}
\dc{skyblue}{RGB}{0, 102, 204}
\dc{jade}{rgb}{0.0, 0.66, 0.42}
\nc{\red}[1]{\textcolor{red}{#1}}
\nc{\blue}[1]{\textcolor{blue}{#1}}
\nc{\green}[1]{\textcolor{green}{#1}}
\nc{\purple}[1]{\textcolor{purple}{#1}}
\nc{\bI}{\bm{\mathrm{I}}}
\nc{\violet}[1]{\textcolor{violet}{#1}}
\nc{\wpink}[1]{\textcolor{watpink}{#1}}
\nc{\jade}[1]{\textcolor{jade}{#1}}
\nc{\watyellow}[1]{\textcolor{watyellow}{#1}}
\nc{\gsublue}[1]{\textcolor{gsublue}{#1}}

\nc{\yck}[1]{\wpink{[YCK: #1]}}
\nc{\hyt}[1]{\blue{[HYT: #1]}}
\nc{\zz}[1]{\watyellow{[ZZ: #1]}}
\nc{\dl}[1]{\green{[DL: #1]}}
\nc{\yckc}[1]{\gsublue{[YCK to myself: #1]}}

\nc{\R}{\mathbb R}
\nc{\D}{\mathcal D}
\nc{\Z}{\mathcal Z}
\nc{\ip}[2]{\langle#1,#2\rangle}
\nc{\zbar}{\overline z }
\nc{\ebar}{\overline e}

\usepackage[textsize=tiny]{todonotes}

\nc{\lei}[1]{%
  \todo[linecolor=blue,backgroundcolor=blue!25,bordercolor=blue]{Lei: #1}%
}
\nc{\wy}[1]{%
  \todo[linecolor=red,backgroundcolor=red!25,bordercolor=red
  ]{Weiyi: #1}
}
\nc{\zl}[1]{%
  \todo[linecolor=green!50!black,backgroundcolor=green!25,bordercolor=green!50!black]{Zilong: #1}%
}
\nc{\yt}[1]{%
  \todo[linecolor=orange!80!black,backgroundcolor=orange!25,bordercolor=orange!80!black]{Yi-Ting: #1}%
}
\nc{\ck}[1]{%
  \todo[
    linecolor=violet!80!black,
    backgroundcolor=violet!25,
    bordercolor=violet!80!black
  ]{Chi-Kuang: #1}}%

\nc{\jx}[1]{%
  \todo[
    linecolor=cyan!70!black,
    backgroundcolor=cyan!20,
    bordercolor=cyan!70!black
  ]{Junxi: #1}%
}

\usepackage{hyperref}
\usepackage{amsmath}
\usepackage{amssymb}
\usepackage{mathtools}
\usepackage{amsthm}

\usepackage[capitalize,noabbrev]{cleveref}

\theoremstyle{plain}
\newtheorem{theorem}{Theorem}[section]

\newtheorem{lemma}[theorem]{Lemma}

\theoremstyle{definition}

\newtheorem{assumption}[theorem]{Assumption}
\theoremstyle{remark}

\usepackage[textsize=tiny]{todonotes}

\usepackage{tikz}
\usetikzlibrary{positioning, shapes.geometric, arrows.meta}
\usepackage{xspace}

\definecolor{highlightpurplebg}{RGB}{238,230,250}
\definecolor{highlightgreenbg}{RGB}{226,245,232}

\definecolor{methodorange}{RGB}{180,85,20}
\newcommand{\pumethod}{%
  \textcolor{methodorange}{\textsc{PUAudit}}\xspace
}

\newcommand{\hlgreen}[1]{%
  \begingroup
  \setlength{\fboxsep}{1.2pt}%
  \colorbox{highlightgreenbg}{#1}%
  \endgroup
}

\definecolor{highlightredbg}{RGB}{255,205,205}

\newcommand{\hlred}[1]{%

  \begingroup

  \setlength{\fboxsep}{1.2pt}%

  \colorbox{highlightredbg}{#1}%

  \endgroup

}

\title{Quantifying and Auditing LLM Evaluation via Positive–Unlabeled Learning}

\author{%
  Zilong Zhang\textsuperscript{*} \\
  Department of Mathematics and Statistics \\
  Georgia State University \\
  \And
  Yi-Ting Hung\textsuperscript{*}\\
  Department of Mathematics and Statistics \\
  Georgia State University \\
  \AND
  Lei Ding \\
  Department of Statistics\\
  University of Manitoba\\\texttt{lei.ding@umanitoba.ca} \\
  \And
  Chi-Kuang Yeh \\
  Department of Mathematics and Statistics \\
  Georgia State University \\
  \texttt{cyeh@gsu.edu}
}

\begin{document}

\maketitle

\begin{abstract}
  Large Language Models (LLMs) are increasingly used as judges for scalable evaluation, yet such LLM-as-a-Judge systems exhibit systematic biases that are decoupled from semantic quality, most notably verbosity bias. Meanwhile, human supervision is costly and typically selective, yielding reliable positive judgments but leaving most outputs unlabelled and potentially mixed in quality. We formulate LLM evaluation under selective human supervision as a positive–unlabelled learning problem and propose a geometric auditing framework based on Partial Optimal Transport. By aligning a small set of human-verified positives with a reliable subset of unlabelled outputs in a fixed embedding space, our method identifies human-consistent preferences and corrects biased judges without retraining. Experiments demonstrate improved alignment with human preferences, increased robustness to presentation biases, and interpretable confidence estimates, offering a scalable and statistically grounded alternative to existing LLM-as-a-judge pipelines.
\end{abstract}

\section{Introduction}
\label{sec:intro}

Reliable evaluation is a foundational requirement for deploying Large Language Models (LLMs) in real-world settings~\citep{liang2022holistic,gu2024survey}.
In many applications, LLM outputs cannot be judged as simply correct or incorrect.
Instead, quality depends on multiple factors, such as correctness, completeness, and clarity, and often varies across tasks and contexts~\citep{ribeiro2020beyond}.
As a consequence, modern evaluation pipelines struggle to reduce performance to a single objective score. Because of this ambiguity, high-quality evaluation still leans on human supervision.
Yet human annotation is costly, and therefore difficult to scale to the volume of generations produced in practice~\citep{dong2022survey,liu2023geval}.
Human review typically covers only a small slice of model outputs, creating a persistent bottleneck.
To relieve this bottleneck, automated evaluation using an LLM judge has quickly become a dominant paradigm~\citep{verga2024replacing}.
Across widely used benchmarks, LLM judges can correlate with human preferences, offering an attractive trade-off between scalability and fidelity~\citep{zheng2024judging,chiang:2024:chatbot-arena}.
Accordingly, LLM judges are now routinely embedded as core components of evaluation and model-selection pipelines.


Despite this practical success, a critical concern remains: LLM judges exhibit systematic biases that are decoupled from underlying semantic quality \citep{ye2024justice}. Figure~\ref{fig:toy_example} in Appendix~\ref{app:notation} illustrates this phenomenon, where a human-preferred response can be overturned by an LLM judge after a small, content-preserving perturbation to its competitor.
Rather than random noise, such biases reflect stable preferences for superficial input features. Recent work demonstrates this behavior through controlled perturbations and bias attacks, showing that minor variations in presentation or style can reliably alter judgments without changing content~\citep{wang2024large,wu2025style,he2025impact}. Consequently, these biases induce predictable evaluation distortions, such as favoring longer or more stylized responses even when content quality remains constant.

The challenge is further compounded by the manner in which human supervision is collected.
Due to limited annotation budgets, human review is typically selective rather than exhaustive~\citep{liang2022holistic,dubois2023alpacafarm}.
Annotations typically focus on fluent or high-impact outputs, while weaker or ambiguous generations are often left unreviewed. This creates a regime in which negative labels are structurally missing~\citep{bekker2020learning,bekker2019beyond}.
Human-verified examples can be regarded as reliable positives, whereas the unverified pool constitutes a heterogeneous mixture of both latent high-quality and low-quality responses.
Under such selective supervision, standard supervised learning assumptions break down, and naively treating all unlabeled data as negative introduces severe bias~\citep{du2014analysis,kato2019learning}.

In this work, we propose a \emph{Positive-Unlabelled Learning Auditing for LLM-as-Judge} (\pumethod), a lightweight auditing framework that reframes biased LLM evaluation under selective human supervision as a positive and unlabelled (PU) auditing problem \citep{elkan2008learning}.
We posit that the central task is to disentangle the latent quality of a response from the biased evaluation mechanism using only limited human verifications rather than recovering an intrinsic notion of response quality.
To this end, we propose a geometric auditing method based on partial Optimal Transport (POT)~\citep{cuturi2013sinkhorn,chapel2020partial}. Our approach operates entirely in a fixed representation space, without retraining the judge or the underlying models, and seeks to align the distribution of verified positives with a reliable subset of the unlabelled pool. Unlike standard OT, which forces all probability mass to be matched, POT explicitly allows only a fraction of mass to be transported. This property naturally accommodates contamination in the unlabelled set, preventing low-quality responses from being spuriously aligned with verified positives. As a result, POT enables the identification of plausible latent positives while remaining robust to noise and mismatch in the unlabelled distribution.
Overall, \pumethod shifts scalable LLM evaluation from a fully supervised labeling problem to a lightweight, representation-fixed auditing problem. This yields a training-free procedure for recovering responses that are likely to be human-preferred from an unlabelled pool contaminated by low-quality generations.

Our contributions are summarized as follows:
\vspace{-0.3cm}
\begin{itemize}
    \item \textbf{Problem formulation and auditing viewpoint.}
    We formalize LLM evaluation under scarce and selective human supervision, demonstrate why standard supervised assumptions fail in this setting, and recast evaluation bias as a PU auditing problem.

    \item \textbf{A lightweight geometric auditing method.}
    We introduce a POT-based auditing approach that operates on fixed embeddings to identify latent human-consistent preferences from unlabelled data, requiring no retraining or modification of the LLM judge.

    \item \textbf{Empirical effectiveness and robustness.}
    Through extensive experiments across multiple benchmarks and bias scenarios, we show that our method substantially improves alignment with human preferences compared to the original LLM-judge decisions.
    \vspace{-5mm}
\end{itemize}

\section{Related works}
\label{sec:background}

\paragraph{Automated Evaluation Methodologies.}
Current automated evaluation approaches largely fall into two categories: reference-based metrics and reference-free model-based evaluation. Traditional metrics like BLEU~\citep{papineni2002bleu} and ROUGE~\citep{lin2004rouge} rely on lexical overlap with ground truth, which often correlates poorly with human judgment on open-ended generation tasks~\citep{liu2016not, celikyilmaz2020evaluation, zhang2019bertscore}.
Consequently, recent research has shifted toward \emph{LLM-as-a-Judge}, where strong models (e.g., GPT-4) serve as evaluators \citep{liu2023geval, zheng2024judging, gilardi2023chatgpt}.
Approaches typically employ either single-score grading or pairwise comparison protocols~\citep{chiang:2024:chatbot-arena, dubois2023alpacafarm, kim2023prometheus, kim2024prometheus} to approximate human preference.
While these frameworks provide scalability, they treat the judge as a fixed oracle. In contrast, our work treats the judge as a biased instrument requiring external auditing.

\begin{figure*}[t]
    \centering 
    \includegraphics[width=0.95\linewidth]{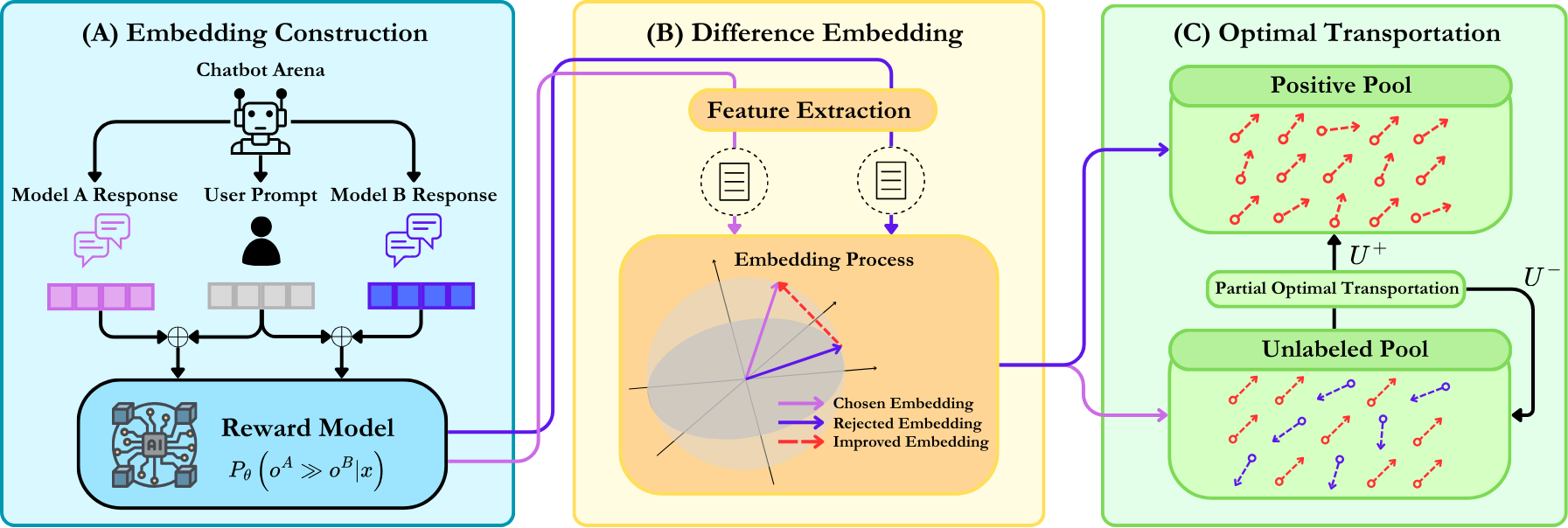}
    \caption{Overview of proposed \pumethod framework; (A) Embedding construction, where pairwise comparison from datasets like Chatbot Arena are processed via a reward model to derive score embedding; (B) Difference Embedding, which extracts feature representations to construct difference vectors that capture the specific direction of quality improvement between winner and loser response;  (C) Optimal Transportation, where POT aligns a clean positive pool of verified improvements with the unlabeled pool.}        \label{fig:framework}
    \label{fig:fullwidth_image}
\end{figure*}

\paragraph{Bias Analysis and Calibration in LLMs.}
Prior work has extensively documented systematic artifacts in LLM judges. Common failures include position bias, where judges favor the first response~\citep{wang2024large, zheng2024judging}; verbosity bias, where longer but vacuous responses are preferred~\citep{saito2023verbosity, singhal2023long}; and self-preference, where models favor their own outputs~\citep{panickssery2024llm}. To address these issues, standard practices include swapping candidate positions to average out order effects~\citep{wang2024large,zheng2024judging},
employing majority voting or juries to reduce variance~\citep{verga2024replacing,cohen2023lm}, or using calibration prompts to enforce neutrality~\citep{zhu2023judgelm}. While such methods mitigate surface-level inconsistencies, they overlook the fundamental distributional shift arising from selective supervision. In contrast, we propose a geometric intervention based on OT to directly align judge outputs with the latent valid distribution.

\paragraph{PU Learning under Selection Bias.}
Our formulation aligns with PU learning, where classifiers are trained using only positive and unlabelled data~\citep{elkan2008learning, du2014analysis,kiryo2017positive,bekker2020learning}. Standard PU methods typically rely on the selected completely at random mechanism, treating labelled examples as an i.i.d. subset of the positive class. Recent advances have relaxed this to instance-dependent selection which is commonly referred as selected at random to handle biased labelling in domains like medical diagnosis or recommendation~\citep{bekker2019beyond, kato2019learning, zamzam:2025:pu-disentangled}.
While existing PU frameworks focus on binary classification or risk estimation, we instead extend this perspective to evaluation auditing. Specifically, we adapt PU principles to disentangle response quality from verification bias, employing POT for robust geometric alignment~\citep{courty2016optimal, chapel2020partial}.

\section{Methodology}\label{sec:method}

\subsection{Problem formulation and notation}\label{sec:formulation}
The framework of \pumethod is shown in Algorithm~\ref{alg:pu_method_pipeline} in Appendix~\ref{subsec:algorithm}. We study the problem of evaluating LLM outputs where reliable human annotations are scarce. In practice, human preferences are selectively available and concentrate on a small subset of samples, while the remaining data are unlabelled and potentially mixed in quality. We formulate this as a PU learning task as follows. Each data pair consists of a user prompt $x_i \in \mathcal X$, where $\mathcal X$ denotes the prompt space, together with two candidate response sequences $o_i^A$ and $o_i^B$ generated by two models, $\mathrm{LLM}_A$ and $\mathrm{LLM}_B$, respectively.
Each pair is associated with preference labels provided by an LLM judge, $\zeta_i^{(\mathrm{LLM})}$, and, when available, a human judge, $\zeta_i^{(H)}$, where $\zeta \in \{\mathrm{A}, \mathrm{B}, \mathrm{Tie}\}$ indicates whether the response from $\mathrm{LLM}_A$ is preferred to that from $\mathrm{LLM}_B$, or whether the two responses are judged as a tie.
In this work, we focus on samples for which a strict preference exists, and exclude ties in the current formulation as the directional difference embedding requires an ordered winner--loser pair. 

Because human preferences are observed only for a subset of samples, we partition the data into annotated and unannotated subsets. The annotated subset contains samples with both human and LLM-judge preferences,
\(
\mathcal{H}
=
\{(x_i, o_i^A, o_i^B, \zeta_i^{(H)}, \zeta_i^{(\mathrm{LLM})})\},
\)
whereas the unannotated subset contains samples with only LLM-judge preferences,
\(
\mathcal{H}^{\prime}
=
\{(x_i, o_i^A, o_i^B, \zeta_i^{(\mathrm{LLM})})\}.
\)
We define an alignment indicator with human preference  
$y_i:= y(\zeta_i) = \mathbb I \left\{\zeta_i = \zeta_i^{(H)}\right\}$ yielding the induced set 
\(\D_P := \{(x_i, o_i^w, o_i^\ell, y_i)\}\), 
where $o_i^w, o_i^\ell \in \mathscr O$ denote the preferred and dispreferred responses, respectively, as determined \textbf{by the human judge $\zeta_i^{(H)}$}. By construction, every sample in $\D_P$ is a verified positive, so $y_i=1$ for all $(x_i, o_i^w,o_i^\ell,y_i)\in\D_P$. For samples in $\mathcal H^\prime$, human preferences are unavailable, and only the LLM-judge preference is observed. The resulting dataset is therefore written as
$
\D_U := \{(x_i, o_i^w, o_i^\ell)\},
$
where the preference ordering $o_i^w, o_i^\ell$ are \textbf{determined by the LLM judge $\zeta_i^{(LLM)}$}. Although a verified preference can be reversed syntactically, the negative class in our problem is not the reversed response pair. It corresponds to cases where the LLM judge's preferred response disagrees with the human preference under the original evaluation protocol. Note that in this set, the latent unobserved $y_i$ can be either zero or one. The summary of main notation used in the paper is shown in Table~\ref{tab:notation} in Appendix~\ref{app:notation}.


Our goal is to learn a scoring function $f:\R^d\to[0,1]$ that estimates $p(y=1\mid o^w,o^\ell)$ for $o^w, o^\ell \in\D_U$ by aligning the unlabelled data with the manifold of $\D_P$. Before formally introducing our method, we define the notation used throughout the paper.  We denote the 
\(\ell_2\)-norm by \(\lVert \cdot \rVert\), the cardinality of a set by \(|\cdot|\), the inner product by \(\langle \cdot, \cdot \rangle\), and distance metrics \(d(\cdot, \cdot)\) where we use cosine similarity in the rest of the manuscript and $\lfloor \cdot \rfloor$ denote the floor function.

\subsection{Feature construction and preprocessing}
\label{subsection:feature}

\paragraph{Fixed representation space.}
Our auditing procedure operates entirely in a fixed representation space. 
Given a prompt $x$ and a response $o$, we use the pretrained encoder $F_\theta$ with fixed parameters to extract a prompt--response representation $F_\theta(x,o)\in\mathbb{R}^d$. 
Importantly, $F_\theta$ is never trained or fine-tuned in our pipeline; all PU relabeling and auditing steps are performed on top of these fixed embeddings. 
In our main experiments, we instantiate $F_\theta$ with a reward-model encoder, since reward models are naturally trained on preference data and provide strong representations for response comparison. 
However, the encoder is modular and can be replaced by any pretrained embedding model. We evaluate several frozen alternatives, including general-purpose embedding models and reward-model encoders. In our pipeline, the Skywork reward-model gives the strongest preference, so we use it as the default encoder. Additional comparisons are reported in Table~\ref{tab:encoder_choice} in Appendix~\ref{app:notation}.

For each pairwise comparison, let $o^w$ and $o^\ell$ denote the response preferred and dispreferred by the initial judge. 
We represent the comparison by the normalized difference embedding
\begin{equation}
    z_i
    =
    \frac{
    F_\theta(x_i,o_i^w)-F_\theta(x_i,o_i^\ell)
    }{
    \|F_\theta(x_i,o_i^w)-F_\theta(x_i,o_i^\ell)\|
    } \in \mathbb{S}^{d-1}.
    \label{eq:z}
\end{equation}
This representation focuses on the direction of relative improvement rather than the absolute embedding of either response. 
It is especially useful when both candidate responses are fluent and semantically close, because the differenced representation directly encodes how the judge separates the preferred response from its counterpart.

\paragraph{Positive group denoising.}
Although human-verified positives provide valuable supervision, they may still contain annotation noise or geometrically inconsistent examples. We therefore apply a lightweight two-stage filter before POT to remove geometric outliers, rather than altering consistent human preference patterns. To avoid conflating semantic topics, this filtering is performed within each question type rather than over the pooled dataset. First, within each question type, we retain positive examples whose embeddings are closest to the corresponding positive-group center. Second, among the retained examples, we further keep comparisons whose improvement directions $z_i$ are most aligned with the average positive improvement direction within the same question type. This produces a cleaner $\D_P'$ that is more geometrically coherent while reducing the risk of discarding valid positives from minority topics. The full method is provided in Appendix~\ref{app:p_denoising}.

\subsection{Partial optimal transport and relabelling}\label{sec:pot}

We align the unlabelled set $\D_U$ to the denoised positive reference set $\D_P^\prime$ using POT. Unlike standard OT, which consider all the instances to transport to another which in our case may transport the unwanted negative instance that the LLM labelled incorrect. Instead, POT allows only a fraction $m$ of the total mass from $\D_U$ to be transported, while the remaining mass is left unmatched.
This design reflects that $\D_U$ contains a mixture of positive and negative samples, whereas $\D_P^\prime$ serves as a clean anchor distribution. Let 
$\mu_P^\prime=\sum_{(x_i,o_i^w,o_i^\ell, y_i)\in\D_P^\prime}a_i$ and $\mu_U=\sum_{(x_j,o_j^w,o_j^\ell)\in\D_U}b_j$ be the empirical measures of the positive and unlabeled sets, where the size of $|\D^\prime_P|=n^\prime_P$, weights $a_i=1/n^\prime_P$ and $b_j=1/n_U$. Let the cost $C_{ij}=1-d(z_i,z_j)$ as the cosine distance between the normalized embeddings and we seek a transport plan $\Gamma\in\R^{n^\prime_{{P}}\times n_U}_{\geq 0}$ via the following objective function that minimizes the transport cost while moving a specific total amount of mass $m\in(0,1]$
\begin{equation*}
\Gamma^\star
= \arg\min_{\Gamma\in\mathbb{R}_{\geq 0}^{n_P^\prime\times n_U}}
\sum_{(x_i,o_i^w,o_i^\ell,y_i)\in \D_P^\prime}
 \sum_{(x_j,o_j^w,o_j^\ell)\in \D_U}
 C_{ij}\Gamma_{ij} \quad
\text{s.t.}\quad 
\Gamma\mathbf{1}_{n_U}\leq \mathbf{a},\quad
  \Gamma^\top\mathbf{1}_{n_P^\prime}\leq \mathbf{b}.
\end{equation*}
The transported mass is further constrained by
\[
\sum_{(x_i,o_i^w,o_i^\ell,y_i)\in \D_P^\prime}
\sum_{(x_j,o_j^w,o_j^\ell)\in \D_U}
\Gamma_{ij}=m .
\]
The inequality constraints allow part of the mass to remain unmatched, avoiding the alignment of geometrically distant negatives in $\D_U$ with the positive anchor distribution $\D_P^\prime$~\citep{chapel2020partial}. 
 For each unlabelled sample $(x_j,o_j^w,o_j^\ell)\in \D_U$, we define an alignment score 
\[
s_j = \sum_{(x_i,o_i^w,o_i^\ell,y_i)\in \D_P^\prime}\Gamma^\star_{ij}~\Rightarrow~\bar s_j = \frac{s_j}{\max_{(x_k,o_k^w,o_k^\ell,y_k)\in \D_U} s_k},
\]
and denote the normalized score as $\bar s_j$. Intuitively, $\bar{s}_j$ measures how strongly the $j$th sample in $\D_U$ aligns with the verified-positive manifold represented by $\D_P^\prime$. If $\bar{s}_j < \eta$, the sample is weakly aligned with the verified-positive manifold, suggesting that the judge preference may be affected by bias. We therefore flip the judge preference $\zeta_j^{(\mathrm{LLM})}$. This gives the corrected judge preference
\[
\hat\zeta_j^{(\mathrm{LLM})} = \left\{ \begin{array}{rcl}
& \mathrm{Flip}(\zeta_j^{(\mathrm{LLM})})  & \mbox{if}~\bar s_j<{\eta},\\ &\zeta_j^{(\mathrm{LLM})} & \mbox{otherwise}, 
\end{array}\right.
\]
where $\eta$ is a calibrated decision threshold with default value \(0.5\) and $\mathrm{Flip}(\cdot)$ returns the opposite preference label. This step targets cases where biases, such as verbosity bias, lead the judge to prefer responses that are geometrically distant from the verified-positive manifold.

\section{Theoretical guarantee}\label{sec:theory}



To rigorously justify our geometric approach, we analyze the consistency of the POT alignment score. Under standard geometric assumptions motivated by the manifold view of representation learning, we show that POT recovers the latent positive distribution and is robust to systematic biases.

\subsection{Geometric setup}
We consider the normalized deviation space on the unit hypersphere $\mathbb S^{d-1}$, where each point $z$ represents a normalized difference between preferred and dispreferred response embeddings defined in Eq.~\eqref{eq:z}. We assume the unlabelled set $\D_U$ is drawn from a mixture distribution $\mu_U=\alpha\mu_++(1-\alpha)\mu_-$ where we denote the distribution of positive and negative/neutral deviations as $\mu_+$ and $\mu_-$, respectively, and $\alpha=(0, 1)$ is the unknown prior probability of human-consistent preference. The positive set $\D_P$ is drawn purely from positive distribution $\mu_P=\mu_+$. 

In our proposal, the human-preferred improvements are not randomly distributed but share common semantic directions, whereas low-quality deviations are more isotropic or clustered differently. To analyze the behaviour of POT on these embeddings, we introduce the following geometric assumptions grounded in the manifold hypothesis of representation learning~\citep{wang2020understanding}. The human-preferred improvement are not isotropic while errors are orthogonal to quality.
\begin{assumption}[Directional Concentration of Improvements]\label{Assumption:direction} The positive measure $\mu_+$ is supported on a cone $\mathcal{C}_\tau$ centred at a consensus improvement direction $\zbar\in\mathbb
S^{d-1}$ with concentration parameter $\tau\in(0,1)$ such that $\text{supp}(\mu_+)\subseteq\mathcal{C}_\tau(\zbar):=\{z\in\mathbb{S}^{d-1}:d(z,\zbar)\geq \tau\}$.

\end{assumption}

\begin{assumption}[Separability of Noise]\label{assumption:separate-noise}
    Let $\tau\in(0,1)$ be the concentration parameter of the improvement cone $\mathcal{C}_\tau(\bar z)$. There exists a safety margin $\delta>0$ such that for any negative deviation $z^-\in\text{supp}(\mu_-)$, the projection onto the consensus direction satisfies $\langle z^-,\bar z\rangle<\tau-\delta$. We further require that $\delta$ is sufficiently large such that $\tau^2>\tau-\delta$ to ensure a non-empty utility gap.

\end{assumption}

This assumption is justified by the improvement-alignment step in the denoising phase, which filters samples based on their cosine alignment and encourages $\D_P$ to concentrate around a consistent direction. The contrastive learning theories also prove that the semantically similar pairs align on the hypersphere~\citep{wang2020understanding}. In addition, the margin $\delta$ ensures that the positive support forms a disjoint region in the embedding space which is distinguishable via cosine similarity.

\subsection{Algorithm guarantee: recovery and robustness}
We now formalize how the latent quality structure within these datasets. To convert the abstract auditing problem into a tractable geometric task, we utilize the latent representations of a reward model to construct difference embeddings. By encoding the vector difference, we transform discrete comparison pairs into a normalized manifold that explicitly captures the gradient of quality improvement, thereby facilitating the subsequent alignment via POT.
\begin{theorem}[High probability recovery of POT]\label{thm: recovery}
Given Assumptions~\ref{Assumption:direction} and~\ref{assumption:separate-noise}, consider the POT problem between $\mu_P$ and $\mu_U$ with cosine cost $C_{ij}=1-d(z_i,z_j)$ and the transport mass $m$. If $m\leq\alpha$ where $\alpha$ is a constant, then as the labelled positive sample size $n_P\to \infty$, the POT plan $\Gamma^\star$ minimizes the transport cost by matching verified positive samples in $\D_P$ to hidden positive samples. Thus,
    \[
    \text{supp}(\Gamma^*)\subseteq\text{supp}(\mu_P)\times \text{supp}(\mu_+),
    \]
and the hidden positive subset $\D_U^+$ in the unlabelled pool can be recovered by POT. Consequently, the POT alignment score is non-zero if and only if $z\in \D_U^+$.
\end{theorem}
The detailed proof can be found in Appendix~\ref{appendix:theorem-recovery}. The key challenge in LLM-as-a-Judge is susceptibility to superficial biases (e.g., verbosity), which we model as nuisance variables. We formally assume these biases are distinct from quality.
\begin{assumption}
    [Latent approximate orthogonality in high-dimensional space]\label{assumption:orthogonal}
    We assume the fixed representation space $\mathbb{R}^d$ encoded by the pre-trained reward model is of sufficiently high dimension that the latent quality direction $\zbar$ and a nuisance direction $z_{bias}$ are approximately orthogonal. Therefore, we assume $\langle \zbar,z_{bias}\rangle\approx0$ without further requirement.
\end{assumption}
\begin{theorem}[Stability under perturbation]\label{thm: robust}
    Given the Assumption~\ref{assumption:orthogonal}, when evaluating the pairwise transport cost between an empirical anchor $z_i\in\D_p^\prime$ and a perturbed embedding $\tilde{z}$, the similarity introduces a first-order sensitivity due to the natural deviation of $z_i$ from $\zbar$. However, because of selective anchors that are highly aligned with $\zbar$, the magnitude of the orthogonal component $\|z_i^\perp\|$ is bounded by a small threshold $\epsilon_p$. Thus, the transport cost is invariant to the bias up to $\mathcal{O}(\lambda\cdot\epsilon_p)+\mathcal{O}(\lambda^2).$ This establishes that while empirical robustness is not purely second-order, the first-order vulnerability is tightly controlled by the denoising phase.
\end{theorem} 
The detailed proof is provided in Appendix~\ref{appendix:theorem-robustness}. This explains why our proposed method outperforms standard judges on the attack (Figure~\ref{fig:attack_result}). While a linear classifier might be unstable by the magnitude of $\lambda z_{bias}$, cosine similarity projects this noise onto the tangent, where its effect vanishes. Additionally, the cosine similarity is involved in the later POT that the ignorable bias would not affect its robustness. We further formalize the limits of this robustness.

\section{Experimental setup}
\label{sec:experiments}

\subsection{Real-data construction}
\label{sec:real_data_construction}

Our experiments are built on the \textit{Chatbot Arena} corpus \citep{zheng2024judging,chiang:2024:chatbot-arena} and \textit{MT-Bench}\citep{zheng2024judging}, which provides pairwise comparisons between two candidate responses together with human preference annotations.
From this source, we first obtain a human-annotated dataset $\tilde{\mathcal{H}}
=\{(x_i, o_i^A, o_i^B, \zeta_i^{(H)})\}$. To obtain LLM judgments, we apply locally deployed LLMs to $\zeta_i^{(\mathrm{LLM})}$.
Using the judge prompt shown in Figure~\ref{fig:judge_prompt} in Appendix~\ref{supp:prompt}, we apply locally deployed LLMs to produce a corresponding preference label $\zeta_i^{(\mathrm{LLM})}$.
This yields the final dataset used throughout our experiments: $\mathcal{H}
=
\{(x_i, o_i^A, o_i^B, \zeta_i^{(H)}, \zeta_i^{(\mathrm{LLM})})\}$. The resulting data naturally induces a PU structure: data with reliable human verification form the positive set, while the remaining data, for which only $\zeta^{(\mathrm{LLM})}$ is available, constitute an unlabeled pool that may contain both correct judgments and latent errors. In real data, the fraction of positives within the unlabeled set is unknown. We therefore estimate the POT fraction $m$ used in Section~\ref{sec:pot} directly from the positive set.
In specific, we hold out a portion of the human-annotated data and measure the consistency between  $\zeta^{(H)}$ and $\zeta^{(\mathrm{LLM})}$.
This empirical consistency is used as a proxy for the positive mass in the unlabeled pool and serves to estimate the transport fraction $m$. We evaluate three locally deployed open-weight models:
\textsc{Mistral-7B-Instruct}~\citep{jiang2023mistral7b},
\textsc{Qwen2.5-7B-Instruct}~\citep{qwen25_technical_report}, and
\textsc{Llemma-7B-MuInstruct}~\citep{azerbayev2023llemma},
together with a stronger closed-source frontier judge, \textsc{GPT-5.4-mini}~\citep{openai_gpt54_blog}.
All judges share the same prompt template. The overall data construction procedure is summarized in Algorithm~\ref{alg:data_framework} in Appendix~\ref{subsec:algorithm}.

\subsection{Robustness evaluation via attack--judge protocol}
\label{sec:attack_judge}

We adopt an bias attack protocol to evaluate the robustness of \pumethod under biases.
Rather than assessing whether the underlying LLM judge itself is robust, our objective is to examine whether \pumethod can effectively mitigate the impact of such biases on downstream evaluation outcomes. Our choice of attacks is informed by \citet{ye2024justice}. They identify 12 common sources of bias affecting LLM-as-judge systems, including verbosity bias, sentiment bias, and distraction effects.
Following this taxonomy, we focus on a representative subset of attacks that are prevalent in practice.

Concretely, starting from the subset $H^\prime$ of pairwise comparisons for which only the original LLM-judge preference $\xi_i^{\text{LLM}}$ is available, we apply controlled perturbations to the candidate responses while preserving their underlying semantic content. We then re-apply the same offline judge to the perturbed response pairs to obtain attacked judgments $\xi_i^{\text{LLM}}$. Comparing $\xi_i^{\text{LLM}}$ and $\tilde\xi_i^{\text{LLM}}$ isolates instabilities induced purely by presentation bias.

\textbf{Length, Sentiment, and Distraction Attacks.} To probe robustness against common presentation-level artifacts, we consider three representative perturbations: length, sentiment, and distraction attacks.
Each attack targets a different superficial aspect of response presentation while preserving the core semantic content. The \emph{length attack} modifies response verbosity to assess sensitivity to excessively long or short answers and to control for prompt budget effects.
The \emph{sentiment attack} alters the emotional tone of responses, testing whether judgments are influenced by affective cues rather than substantive quality.
The \emph{distraction attack} injects irrelevant or misleading textual elements, evaluating robustness to superficial noise. All three attacks follow a unified attack--judge protocol and differ only in how the surface form of candidate responses is perturbed.
The corresponding prompt templates are provided in Figures~\ref{fig:attack_distraction_position}, \ref{fig:attack_length} and \ref{fig:attack_sentiment} in Appendix~\ref{supp:prompt}. 

\begin{table*}[h]
\centering
\caption{Original and adjusted consistency ratios across question types. Each cell reports the mean value (standard deviation) over 10 random seeds from repeated random splits of the human-annotated data. Improvement is computed as the difference between the adjusted and original ratios. For each model block, the question-type sample size is calculated as the sum of the positive (P) and unlabelled (U) groups. Boldface values indicate positive improvements achieved by our method. Entries marked with * indicate extremely small sample sizes and are excluded from further analysis. Definitions of each question type are provided in Appendix~\ref{fig:judge_prompt}.}
\resizebox{\textwidth}{!}{%
\begin{tabular}{l|cccccc}
\toprule
\textbf{Model: Mistral}
& \textbf{QTA (n=7660)} & \textbf{QTB (n=88)} & \textbf{QTC (n=8124)}
& \textbf{QTD (n=1294)} & \textbf{QTE (n=1039)} & \textbf{QTF (n=4998)} \\
\midrule
Orig. ratio
& 58.41\% (4.45e-03) & 50.32\% (3.02e-02) & 52.97\% (3.32e-03)
& 59.08\% (9.42e-03) & 50.19\% (6.12e-03) & 37.56\% (4.27e-03) \\
Adj. ratio
& 67.90\% (4.65e-03) & 62.90\% (5.32e-02) & 66.73\% (5.51e-03)
& 67.25\% (1.08e-02) & 64.62\% (1.53e-02) & 63.47\% (6.49e-03) \\
Improvement
& \hlgreen{\textbf{9.50\%}} & \hlgreen{\textbf{12.58\%}} & \hlgreen{\textbf{13.76\%}} & \hlgreen{\textbf{8.17\%}} & \hlgreen{\textbf{14.43\%}} & \hlgreen{\textbf{25.91\%}} \\
\midrule
\textbf{Model: Qwen}
& \textbf{QTA (n=4395)} & \textbf{QTB (n=1957)} & \textbf{QTC (n=2105)}
& \textbf{QTD (n=514)} & \textbf{QTE (n=8472)} & \textbf{QTF (n=128)} \\
\midrule
Orig. ratio
& 67.57\% (5.73e-03) & 66.54\% (5.54e-03) & 68.64\% (6.68e-03)
& 68.49\% (1.62e-02) & 63.53\% (5.63e-03) & 56.37\% (3.26e-02) \\
Adj. ratio
& 70.01\% (3.51e-03) & 70.91\% (7.60e-03) & 70.18\% (1.04e-02)
& 67.79\% (2.24e-02) & 63.42\% (4.88e-03) & 63.29\% (3.91e-02) \\
Improvement
& \hlgreen{\textbf{2.44\%}} & \hlgreen{\textbf{4.37\%}} & \hlgreen{\textbf{1.54\%}} & \hlred{-0.70\%} & \hlred{-0.12\%} & \hlgreen{\textbf{6.92\%}} \\

\midrule
\textbf{Model: Llemma}
& \textbf{QTA (n=525)} & \textbf{QTB (n=172)} & \textbf{QTC (n=843)} & \textbf{QTD (n=10)*} & \textbf{QTE (n=2)*} & \textbf{QTF (n=3)*} \\
\midrule
Orig. ratio
& 51.47\% (1.85e-02) & 64.16\% (2.07e-02) & 55.94\% (7.49e-03) & - & - & - \\
Adj. ratio
& 54.67\% (2.67e-02) & 52.27\% (7.68e-02) & 58.45\% (1.23e-02) & - & - & - \\
Improvement
& \hlgreen{\textbf{3.20\%}} & \hlred{-11.88\%} & \hlgreen{\textbf{2.51\%}} & - & - & -  \\
\bottomrule
\end{tabular}}
\label{tab:sample}
\end{table*}

\section{Empirical evaluation}
\label{sec:results}

We evaluate \pumethod via ($\mathrm{Cons}_{orig},\mathrm{Cons}_{adj}$) defined in Appendix~\ref{sec:metrics} from two complementary perspectives:
(i) whether the \pumethod improves alignment with limited human judgments and
(ii) whether such improvements persist under systematic presentation biases.
Across all experiments, we report consistency ratios across question types, with detailed breakdowns by model and attack.

\subsection{Main result: alignment and robustness}
\label{sec:results:clean}

We begin with the no-attack setting, which serves as a baseline for assessing whether \pumethod improves agreement with human preferences. Table~\ref{tab:sample} reports the original consistency ratio ($\mathrm{Cons}_{\mathrm{orig}}$) and the adjusted consistency ratio ($\mathrm{Cons}_{\mathrm{adj}}$) across six question types (QTA--QTF) for two judges, Mistral and Qwen; the full results are provided in Tables~\ref{tab:none_attack_by_model_qtype_reheader} and \ref{tab:mtbench} in Appendix~\ref{app:mian result}.

Overall, \pumethod yields systematic improvements across question types and LLM judges. For Mistral, $\mathrm{Cons}_{\mathrm{adj}}$ exceeds $\mathrm{Cons}_{\mathrm{orig}}$ for all six question types, with QTF (Other) being less stable due to its limited sample size. For Qwen, the gains are more modest but remain broadly consistent: four of the six question types improve, while Advise (QTD) and Writing (QTE) drop by less than one percentage point, suggesting that \pumethod largely preserves already well-aligned judge behavior.

\begin{figure*}[tb]
    \centering
    \includegraphics[width=0.9\textwidth]{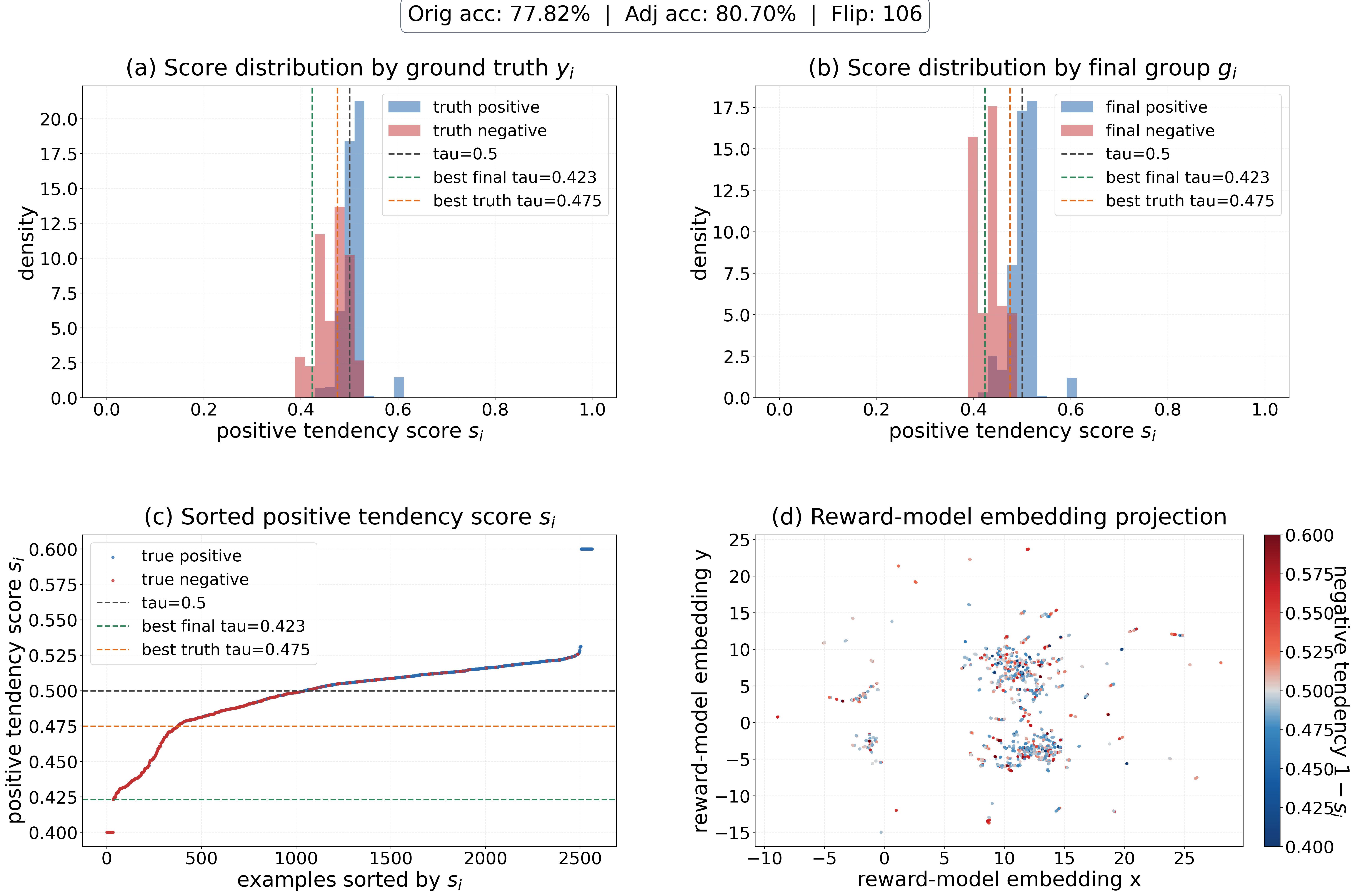}
    \caption{Diagnostic visualization of the POT-based alignment score on the MT-Bench run. Panels (a) and (b) compare score distributions by ground-truth consistency label \(y_i\) and final group assignment \(g_i\), respectively. Panel (c) sorts examples by the diagnostic score \(s_i=0.6(1-r_i)+0.4(1-d_i)\), with reference thresholds for the default rule, final groups, and ground truth. Panel (d) shows a two-dimensional projection of the reward-model embeddings colored by negative tendency \(1-s_i\). 
The summary box reports original accuracy, adjusted accuracy, and the number of flipped examples.}
    \label{fig:tendency}
\end{figure*}

Figure~\ref{fig:tendency} illustrates how \pumethod modifies decisions on MT-Bench using the final positive-tendency score, i.e., the refined responsibility that the original LLM-judge decision is human-consistent. The top panels show density estimates of this score, the bottom-left panel sorts examples by score with the final assignment thresholds, and the bottom-right panel projects fixed reward-model embeddings into two dimensions and colors points by the final score. The isolated blue point in the upper-right of the sorted-score panel is a high-score example assigned to the final positive group.

\begin{figure*}[tb]
    \centering
    \includegraphics[width=1\textwidth]{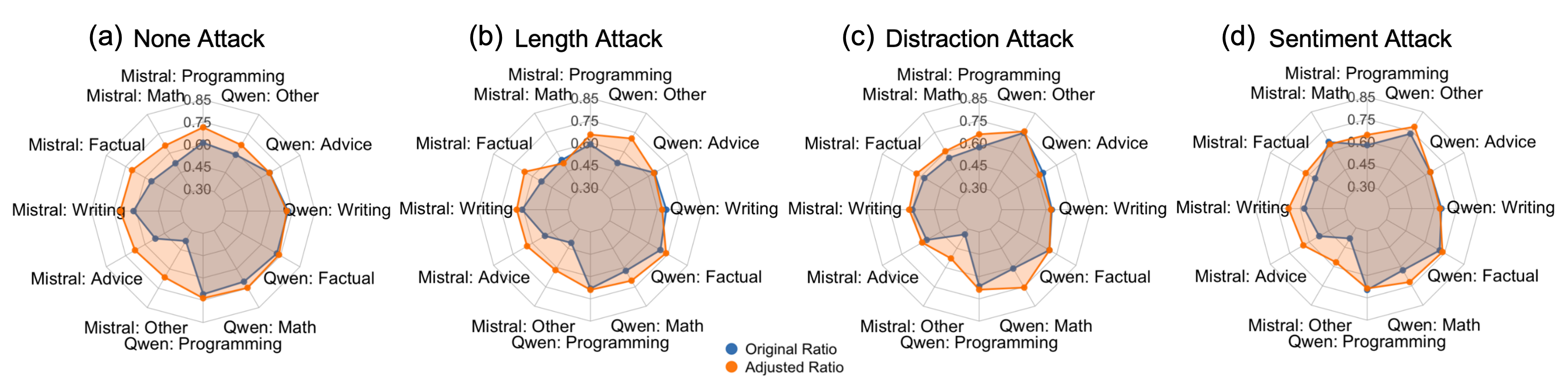}
    \caption{Comparison of original and adjusted consistency ratios across question types and models. The blue region indicates the original consistency ratio, while the orange region indicates the adjusted consistency ratio. Improvements after adjustment are reflected by the adjusted region extending beyond the original one.}
    \label{fig:attack_result}
\end{figure*}

We further examine the behavior of the method across different attack settings and question types. We evaluate robustness under three presentation bias attacks: \emph{Length}, \emph{Distraction}, and \emph{Sentiment}, which alter surface features while retaining semantic content. As shown in Figure~\ref{fig:attack_result}, all attacks greatly reduce $\mathrm{Cons}_{\mathrm{orig}}$. \pumethod consistently recovers enormous lost alignment, with $\mathrm{Cons}_{\mathrm{adj}}$ remaining markedly higher than $\mathrm{Cons}_{\mathrm{orig}}$ under all attacks. The radar plots further reveal a qualitative change in evaluation behavior. Before adjustment, consistency varies sharply across question types and judges, producing anisotropic profiles. After the PU alignment, the profiles become heavily more uniform, with radar shapes closer to circular, indicating reduced sensitivity to question type and judge-specific biases. Notably, these gains are gained without access to attacked examples during adjustment.
Aligned with Theorem~\ref{thm: robust}, the results confirm that \pumethod provides an attack-agnostic robustness rather than overfitting to specific perturbations.

According to Figure~\ref{fig:attack_result}, the magnitude of improvement varies across question types.
Programming (QTA) and Math (QTB) show relatively smaller gains, as these tasks are dominated by objective correctness, where LLM judges are already more stable and less sensitive to superficial cues. In contrast, larger improvements are observed for Advice (QTE) and Writing (QTD), which involve open-ended reasoning and subjective quality judgments and are therefore more vulnerable to presentation biases such as verbosity and framing.
This suggests that \pumethod primarily mitigates non-semantic biases rather than interfering with judgments driven by objective correctness. Overall, \pumethod provides the greatest benefit in regimes where LLM-based evaluation is most fragile.

\subsection{Sensitivity analysis}
\label{sec:insensitive}

Finally, we provide a diagnostic sensitivity analysis for two main hyperparameters: the fraction of verified positive examples used during auditing, denoted by $p_{\text{ratio}}\in(0,1]$, and the transported unlabeled mass fraction, denoted by $m\in(0,1]$. The detailed results are reported in Tables~\ref{tab:p_ratio} and ~\ref{tab:transport} in Appendix~\ref{app:mian result}. The results show that performance is not highly sensitive to moderate changes in either parameter. As $p_{\text{ratio}}$ increases, $\mathrm{Cons}_{\mathrm{adj}}$ quickly reaches a plateau, suggesting that a small clean positive set is often sufficient to estimate the main direction of human-consistent refinement. The transport fraction $m$ also affects performance smoothly, with stronger results when it is large enough to cover most latent positives. These trends support the use of \pumethod in limited verification settings, while the full numerical comparisons are deferred to the appendix.

\section{Discussion and future work}\label{sec:conclusion}

This work introduces a PU learning framework that leverages POT and geometric alignment to audit LLM evaluation under scarce and biased human supervision.
Several extensions remain for future study. A primary direction is to remove the current reliance on a specific reward model.
Replacing it with general-purpose contrastive or self-supervised encoders would decouple the framework from particular architectures and improve robustness and transferability across domains.

In addition, our formulation assumes a single direction of quality improvement, whereas response quality is inherently multi-dimensional. Extending the method to support multiple improvement directions or multi-manifold structures would better capture heterogeneous notions of quality and naturally connects to multi-PU or multi-class settings, such as those involving multiple LLM or human judges. Finally, future work may explore alternative distance metrics and alignment objectives beyond cosine similarity and POT, as well as more principled inference of the latent positive proportion in the unlabelled set.

\bibliographystyle{plainnat}
\bibliography{PUaudit}

\appendix
\clearpage
\newpage

\startcontents[appendix]
\section*{Appendix}
\printcontents[appendix]{}{1}{\setcounter{tocdepth}{2}}
\clearpage

\section{Proof}
\label{app:proof}
\subsection{Proof of theorem~\ref{thm: recovery}}
\label{appendix:theorem-recovery}

The goal of POT algorithms is to pair the limited set of verified positive samples with candidates from the unlabelled pool to maximize the total similarity. The unlabelled pool contains latent true positives and latent true negatives.
\paragraph{Primal Formulation.} The POT problem minimizes the total cost $C$ as $\mathcal{L}(\Gamma)$

\begin{align*}
    \mathcal{L}(\Gamma)&=\sum_{(x_i,o_i^w,o_i^\ell,y_i)\in \D_P^\prime} \sum_{(x_j,o_j^w,o_j^\ell)\in \D_U} C_{ij}\Gamma_{ij},\\&\quad\text{subject to }\Gamma\mathbf{1}_{n_U}\leq \mathbf{a},~\Gamma^\top\mathbf{1}_{n^\prime_P}\leq \mathbf{b}, 
    \quad \sum_{(x_i,o_i^w,o_i^\ell,y_i)\in \D_P^\prime}\sum_{(x_j,o_j^w,o_j^\ell)\in \D_U}\Gamma_{ij}=m. 
\end{align*}
Since $C_{ij}=1-d(z_i,z_j)$, the objective is equivalent to maximizing 
\[
\mathcal{J}(\Gamma)=\sum_{(x_i,o_i^w,o_i^\ell,y_i)\in \D_P^\prime} \sum_{(x_j,o_j^w,o_j^\ell)\in \D_U} d(z_i,z_j)\Gamma_{ij}
\]
with the same constraints.

\paragraph{Partitioning and Utility Bounds.} Let $U=U_+~\cup~U_-$, where $U_+=U~\cap~\text{supp}(\mu_+)$ are the latent true positives, and $U_-=U~\cap~\text{supp}(\mu_-)$ are latent true negatives. Let $z_i\in \D_P^\prime$ be an empirical anchor. We decompose any candidate vector $z\in\D_U$ into a parallel and an orthogonal component relative to the consensus direction $\zbar$: $z=\tau_z\zbar+\epsilon_zz^\perp$ where $\tau_z=\langle z,\zbar\rangle,z^\perp$ is a unit vector orthogonal to $\zbar$ and $\epsilon_z=\sqrt{1-\tau_z^2}$.
\begin{itemize}
    \item ($U_+\to P$): Consider a sample $z_j\in U_+$. By Assumption~\ref{Assumption:direction}, both $z_i,z_j$ lies within the improvement cone $\mathcal{C}_\tau$ and $\tau_i,\tau_j>\tau$. Thus, the inner product of these two vectors is
    $$
    \langle z_i, z_j\rangle=\tau_i\tau_j+\epsilon_i\epsilon_j\langle z_i^\perp,z_j^\perp\rangle\geq\tau^2-(1-\tau^2)\max_{j\in U_+}|\langle z_i^\perp,z_j^\perp\rangle|=\tau^2-(1-\tau^2)t
    $$.
    \item ($U_-$ remains): Consider a target $z_k\in U_-$. By Assumption~\ref{assumption:separate-noise}, the negative sample is separated from the consensus direction by a margin $\delta$, meaning $\tau_k\leq \tau-\delta$ where the inner product is bounded by 
    $$
    \langle z_i,z_k\rangle\leq (\tau-\delta)+\max_{k\in U_-}|\langle z_i^\perp,z_k^\perp\rangle|=(\tau-\delta)+t.
    $$
\end{itemize}

Establishing the utility gap between a true positive versus negative that  
\begin{align*}
    \Delta=\min\langle z_i,z_j\rangle-\max\langle z_i,z_k\rangle&\geq(\tau^2-\tau+\delta)-2\max_{x\in U}|\langle z_i^\perp,x^\perp\rangle|\\
    &\geq(\tau^2-\tau+\delta)-2t
\end{align*}
Let the base theoretical gap be defined as $\Delta_{base}=\tau^2-\tau+\delta$. By Assumption~\ref{assumption:separate-noise}, we require $\delta>\tau(1-\tau)$ to ensure $\Delta_{base}>0$. To guarantee $\Delta>0$ we require $2t<\Delta_{base}$. Setting $t=\Delta_{base}/4$, we substitute this into the probabilistic bound established in Lemma~\ref{lemma: lowerbound}
\begin{equation*}
    P\left(\Delta>0\right)\geq1-2N_U\exp\left(-\frac{(d-1)\Delta_{base}^2}{32}\right)
\end{equation*}
As the representation dimension $d\to\infty$, the probability that the utility gap remains strictly positive approaches 1. Therefore, POT will route the transport mass to $U_+$ to minimize the global cose.

\hfill $\blacksquare$

\begin{figure}
    \centering
    \includegraphics[width=0.65\linewidth]{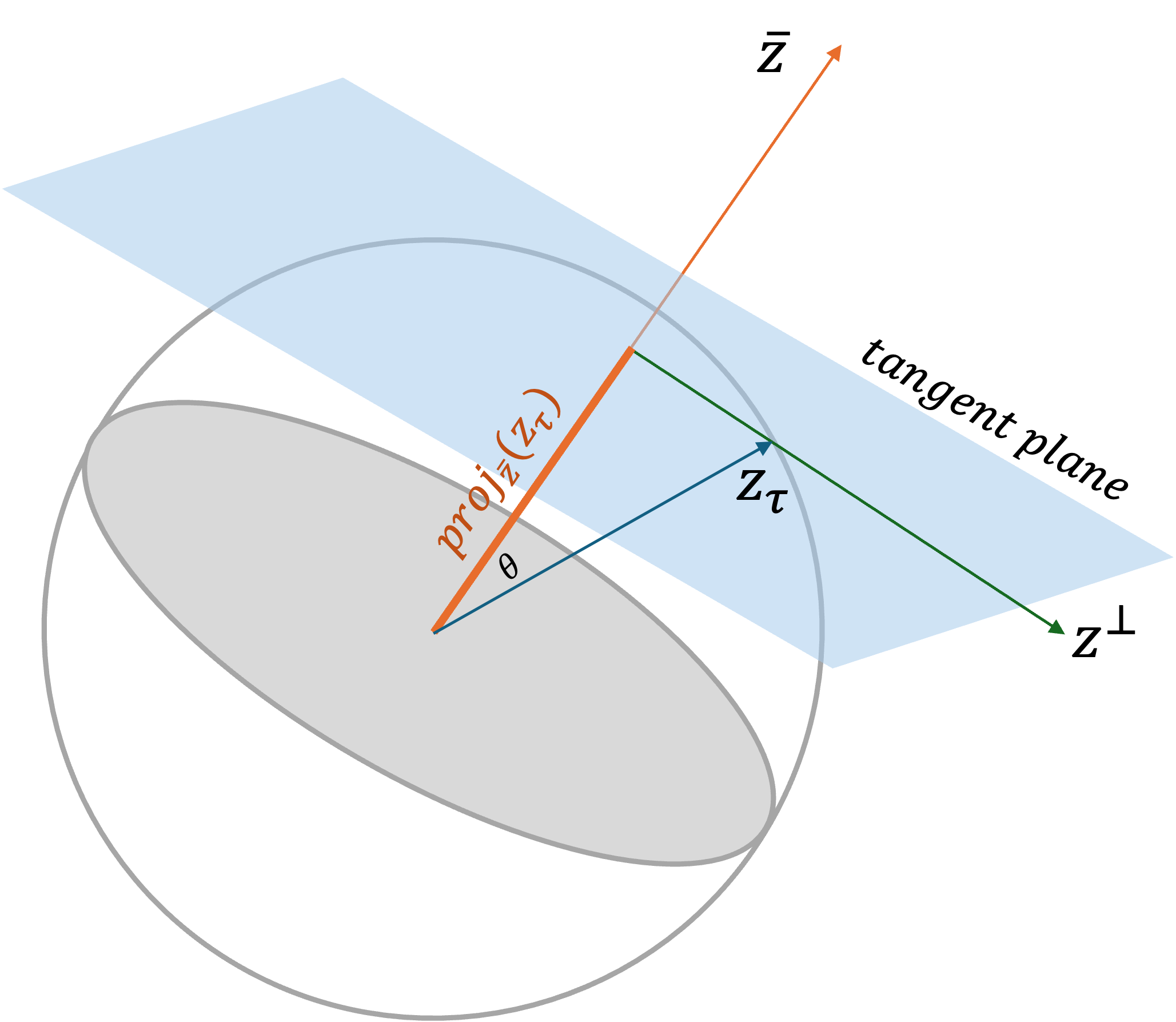}
    \caption{The vector decomposition for Lemma~\ref{lemma: lowerbound}: We denote $\theta$ as the angle between any normalized embedding $z_\tau$ and the consensus direction $\zbar$, and $proj_{\zbar}(z_\tau)$ as the projected amount of $z_\tau$ onto $\zbar$.}
    \label{fig:lemmab1}
\end{figure}

\begin{lemma}\label{lemma: lowerbound}
    Let $\overline{z}$ be the unit consensus direction where we define it as the centre of the positive cone $\mathcal{C}_\tau(\zbar)$ as in Assumption~\ref{Assumption:direction} and let $z_i^\perp,x^\perp$ be unit vectors $(d-1)$-dimensional orthogonal complement subspace (Figure~\ref{fig:lemmab1}). Assuming the unlabeled samples' orthogonal components $\mathcal{X}^\perp=\{x_1^\perp,\dots,x_{N_U}^\perp\}$ are distributed approximately uniformly on the unit sphere of the subspace, the maximum correlation for a fixed anchor $z_i^\perp$ is bounded by $t$ with high probability:
    \begin{equation}
        P\left(\max_{x^\perp\in\mathcal{X}^\perp}|\langle z_i^\perp,x^\perp\rangle|>t\right)\leq N_U\exp\left(-\frac{(d-1)t^2}{2}\right).
    \end{equation}

\paragraph{Proof.}
For any single random unit vector $x^\perp\sim \mathrm{Uniform}(\mathbb{S}^{d-2})$, its projection onto a fixed unit vector $z_i^\perp$ is a sub-Gaussian random variable. By the concentration of measure on the high-dimensional sphere, the tail probability is bounded by 
\begin{equation*}
    P(|\langle z_i^\perp,x^\perp\rangle|>t)\leq 2\exp\left(-\frac{(d-1)t^2}{2}\right).
\end{equation*}
Since the POT algorithm searches over the entire unlabeled pool of size $N_U$, we must bound the worst case maximum inner product to prevent the algorithm from spuriously matching orthogonal noise. Applying the union bound over all $N_U$ samples:

\begin{equation*}
    P(\max_{x^\perp\in\mathcal{X}^{\perp}}|\langle z_i^\perp,x^\perp\rangle|>t)\leq \sum_{x^\perp\in\mathcal{X}^\perp}P(|\langle z_i^\perp,x^\perp\rangle|>t)\leq 2N_U\exp\left(-\frac{(d-1)t^2}{2}\right)
\end{equation*}

This establishes that in high-dimensional space, the maximum orthogonal correlation is strictly bounded, mitigating the risk of POT exploiting spurious correlations.
\end{lemma}

\subsection{Proof of theorem~\ref{thm: robust}}\label{appendix:theorem-robustness}
We imagine an ``Attacker" trying to trick the judge by taking a response $\zbar$ and injecting a bias vector $z_{bias}$ with magnitude $\lambda$.

\textbf{The Attack.} The attacker constructs a perturbed embedding $\tilde{z}$ by adding orthogonal nuisance with magnitude $\lambda$ to $\zbar$. The perturbed target is $\tilde{z}$:
\[
\tilde{z}=\frac{\overline{z}+\lambda z_{bias}}{\sqrt{1+\lambda^2}}.
\]

\textbf{The Defence.} We examine how the similarity between $\tilde{z}$ and $\overline{z}$ changes with the perturbation parameter $\lambda$. The cosine similarity between $z_i$ and $\tilde{z}$ is
\begin{align*}
    d( \overline{z},\tilde{z})
    &=\left\langle z_i, \frac{\overline{z}+\lambda z_{bias}}{\sqrt{1+\lambda^2}} \right\rangle
    =\frac{\langle\tau_i\zbar+\epsilon_iz_i^\perp, \overline{z}+\lambda z_{bias}\rangle}{\sqrt{1+\lambda^2}}\\
    &=\frac{\tau_i+\lambda\epsilon_i\langle z_i^\perp, z_{bias}\rangle}{\sqrt{1+\lambda^2}}\\
    &=\tau_i+\lambda\epsilon_i\langle z_i^\perp,z_{bias}\rangle-\frac{1}{2}\tau_i\lambda^2.
\end{align*}
The change in the alignment score caused by attack is $\lambda\epsilon_i\langle z_i^\perp,z_{bias}\rangle-\frac{1}{2}\tau_i\lambda^2$. The derivation reveals a first-order sensitivity term. However, the \pumethod is designed with a specific geometric intervention to mitigate this, the P-group denoising stage.

By selectively retaining the anchors that exhibit high cosine alignment with the empirical center, the denoising procedure enforces $\tau_i\to1$. Consequently, the magnitude of $\epsilon_i=\sqrt{1-\tau_i}$ is bounded by a small tolerance threshold $\epsilon_p\ll1$.

Since $|\langle z_i^\perp,z_{bias}\rangle\leq 1$, the first-order sensitivity is bounded by $\mathcal{O}(\epsilon_p\cdot\lambda)$. Thus, while the robustness is not purely second-order $\mathcal{O}(\lambda^2)$ as it would be with the theoretical $\zbar$, the first-order disturbance is heavily attenuated by $\epsilon_p$, formalizing the robustness observed against presentation attacks.

\hfill $\blacksquare$

\subsection{Proof of insensitivity to the number of verified positives}
\label{app:p_ratio_theory}

We provide a theoretical explanation for the statement in Section~\ref{sec:insensitive}. The central intuition is that \pumethod utilizes the denoised positive set $\D_P^\prime$ primarily to estimate a stable reference direction (the ``quality vector'') in the difference-embedding space. Once $\D_P^\prime$ reaches a certain cardinality, it provides sufficient information to characterize this direction, rendering additional positive samples redundant.

\begin{assumption}[Directional concentration of $\D_P^\prime$]
\label{ass:p_direction}
There exists a latent unit vector $\zbar$ (representing the gold-standard preference direction) such that for all $(x_i,o_i^w,o_i^\ell,y_i)\in\D_P^\prime$, the corresponding normalized difference embedding $z_i$ satisfies $\langle z_i,\zbar\rangle \;\ge\; 1-\lambda$ for some $\lambda\in(0,1)$ Furthermore, we assume that the latent distribution of these positives is symmetric around $\zbar$ within this spherical cap.
\end{assumption}

\begin{theorem}[Informational Sufficiency of the Positive Set]
\label{thm:p_ratio_sufficient}
Under Assumption~\ref{ass:p_direction}, for any $\varepsilon>0$, there exists a finite threshold $n^\star(\varepsilon, \lambda)$ such that whenever $|\D_P^\prime| \ge n^\star$, the empirical reference direction estimated by \pumethod is $\varepsilon$-aligned with $\zbar$ with high probability.
\end{theorem}

The P-group denoising procedure in Section~\ref{subsection:feature} enforces Appendix~\ref{ass:p_direction}by retaining only co-directional anchors, ensuring that all vectors in $\D_P^\prime$ lie within a narrow spherical cap around $\zbar$. According to the standard concentration of vector averages on the unit sphere, the empirical mean direction converges to the population mean $\zbar$ at a rate of $\mathcal{O}(|\D_P^\prime|^{-1/2})$.

In our framework, $p_{\mathrm{ratio}}$ determines the effective size of $\D_P^\prime$. Once $|\D_P^\prime| \ge n^\star$, the estimated reference direction becomes stable, and further increasing $p_{\mathrm{ratio}}$ yields only marginal gains in variance reduction without providing new structural information. Consequently, the reference anchor for POT remains fixed, leading to the observed plateau in OT-based scores and adjusted consistency $\mathrm{Cons}_{\mathrm{adj}}$. This demonstrates that \pumethod is robust to the quantity of positive data, provided the minimum threshold for representational sufficiency is met.

\subsection{Feature extraction and P-group denoising}
\label{app:p_denoising}
\paragraph{Feature extraction.}
We use a frozen encoder $F_\theta$ to construct the representation space. 
In our main experiments, $F_\theta$ is instantiated with a reward-model encoder, but the method itself does not require a specific encoder architecture. 
For a prompt $x$ and response $o$, we format the input as
\[
\text{input}
=
\texttt{"<|user|>"}
\oplus x
\oplus
\texttt{"<|assistant|>"}
\oplus o,
\]
and use the last-token hidden representation as the prompt--response embedding $F_\theta(x,o)\in\mathbb{R}^d$.

For a pairwise comparison, let $o^w$ and $o^\ell$ be the response preferred and dispreferred by the initial judge. 
We define the comparison representation as
\begin{equation}
    z_i
    =
    \frac{\Delta_i}{\|\Delta_i\|},
    \qquad
    \Delta_i
    =
    F_\theta(x_i,o_i^w)-F_\theta(x_i,o_i^\ell).
\end{equation}
Thus, each comparison is represented by a unit vector on $\mathbb{S}^{d-1}$ indicating the direction from the dispreferred response to the preferred response. 
Compared with absolute response embeddings, this differenced representation directly captures relative preference information and reduces sensitivity to scale variations across examples.

\paragraph{P-core filtering.}
Although human annotations provide valuable supervision, they may still contain noise. 
We therefore first remove positive examples whose preferred-response embeddings are far from the positive-group center. 
For each positive example, let
\[
    e_i = F_\theta(x_i,o_i^w).
\]
We compute the empirical positive center
\[
    \bar e = \frac{1}{|\D_P|}
    \sum_{(x_i,o_i^w,o_i^\ell,y_i)\in \D_P} e_i,
\]
and score each example by cosine similarity
\[
    s_i^{(e)}
    =
    \frac{\langle e_i,\bar e\rangle}{\|e_i\|\,\|\bar e\|}.
\]
We retain the top-$\alpha_1$ fraction of examples according to $s_i^{(e)}$, yielding
\[
    \D_P^{(1)}
    =
    \operatorname{Top}_{\lfloor \alpha_1 |\D_P|\rfloor}
    \left(
    \D_P; s_i^{(e)}
    \right),
\]
where $\alpha_1\in(0,1]$ is a retention ratio.

\paragraph{Improvement-alignment filtering.}
Next, we filter examples according to whether their improvement directions are aligned with the dominant positive direction. 
We compute
\[
    \bar z
    =
    \frac{1}{|\D_P^{(1)}|}
    \sum_{(x_i,o_i^w,o_i^\ell,y_i)\in \D_P^{(1)}} z_i,
\]
and assign each retained example an alignment score
\[
    s_i^{(z)}
    =
    \frac{\langle z_i,\bar z\rangle}{\|z_i\|\,\|\bar z\|}.
\]
The final cleaned positive set is obtained by retaining the top-$\alpha_2$ fraction:
\[
    \D_P'
    =
    \operatorname{Top}_{\lfloor \alpha_2 |\D_P^{(1)}|\rfloor}
    \left(
    \D_P^{(1)}; s_i^{(z)}
    \right),
\]
where $\alpha_2\in(0,1]$ controls the strength of improvement-direction filtering.

In our experiments, we set $\alpha_1=\alpha_2=0.7$. 
This choice removes geometrically inconsistent positive examples while preserving enough positive mass for the subsequent POT step.

\subsection{Evaluation metrics}
\label{sec:metrics}

We evaluate the stability of LLM-as-judge decisions under controlled presentation attacks using a single metric: consistency. 
For a fixed judge model and a fixed attack type, let $\zeta_i^{(\mathrm{H})}$ denote the judge’s preference of human and $\zeta_i^{(\mathrm{LLM})}$ denote the preference of LLM. We define
\begin{equation*}
\mathrm{Cons} \;=\; \frac{1}{|\mathcal{I}|}\sum_{i\in\mathcal{I}} \mathbb{I}\!\left\{\zeta_i^{(\mathrm{LLM})}=\zeta_i^{(\mathrm{H})}\right\},
\end{equation*}
where $\mathcal{I}$ denotes the set of test samples for which both human annotations and LLM judge prediction are available.

We report consistency before and after applying \pumethod. 
Concretely, $\mathrm{Cons}_{\mathrm{orig}}$ is computed using the raw judge outputs $(\zeta_i^{(\mathrm{LLM})},\zeta_i^{(\mathrm{H})})$, while $\mathrm{Cons}_{\mathrm{adj}}$ is computed using the adjusted preferences $(\tilde{\zeta}_i^{(\mathrm{LLM})},\zeta_i^{(\mathrm{H})})$. Higher consistency indicates better alignment between the LLM judge and human judgments, reflecting more stable and human-like evaluation behaviour under presentation-induced perturbations.

\section{Additional plot and table}

\subsection{Notation and setup}
\label{app:notation}

\begin{table}[ht]
\centering
\caption{Main notation used in the paper. Additional symbols are defined locally when introduced.}
\small
\begin{tabular}{ll}
\toprule
\textbf{Symbol} & \textbf{Meaning} \\
\midrule
\multicolumn{2}{c}{\textit{Data and Preference}} \\
\midrule
$\mathcal{D}$ & collection of pairwise comparisons \\
$x_i$ & prompt (question) \\
$o_i^A, o_i^B$ & two candidate responses \\
$\zeta_i^{(\mathrm{LLM})}$ & preference given by the LLM judge \\
$\zeta_i^{(\mathrm{H})}$ & human preference (when available) \\
$o_i^w, o_i^\ell$ & winner and loser induced by the judge \\
$y_i$ & consistency indicator, $\mathbf{1}\{\zeta_i^{(\mathrm{LLM})} = \zeta_i^{(\mathrm{H})}\}$ \\
$\mathcal{D}_P$ & verified positive set (human-consistent) \\
$\mathcal{D}_U$ & unlabeled set (mixture of latent positives and negatives) \\
\midrule
\multicolumn{2}{c}{\textit{Representation}} \\
\midrule
$F_\theta$ & frozen encoder (reward model or embedding model) \\
$z_i$ & normalized difference embedding \\
$\mu_P, \mu_U$ & empirical distributions of $\mathcal{D}_P$ and $\mathcal{D}_U$ \\
$\mu_+, \mu_-$ & latent positive and negative distributions \\
$\alpha$ & latent positive proportion in $\mathcal{D}_U$ \\
\midrule
\multicolumn{2}{c}{\textit{POT and Alignment}} \\
\midrule
$C_{ij}$ & cosine transport cost between embeddings \\
$\Gamma^\star$ & optimal partial transport plan \\
$m$ & transported mass (transport fraction) \\
$s_j$ & transport-based alignment score \\
$\bar{s}_j$ & normalized alignment score \\
$\eta$ & decision threshold for preference correction \\
\midrule
\multicolumn{2}{c}{\textit{Analysis Variables}} \\
\midrule
$p_{\mathrm{ratio}}$ & fraction of verified positives used in auditing \\
$\lambda$ & magnitude of bias perturbation \\
$z_{\mathrm{bias}}$ & nuisance (bias) direction \\
\bottomrule
\end{tabular}
\label{tab:notation}
\end{table}

\begin{figure}[tb]
    \centering
    \includegraphics[width=0.9\textwidth]{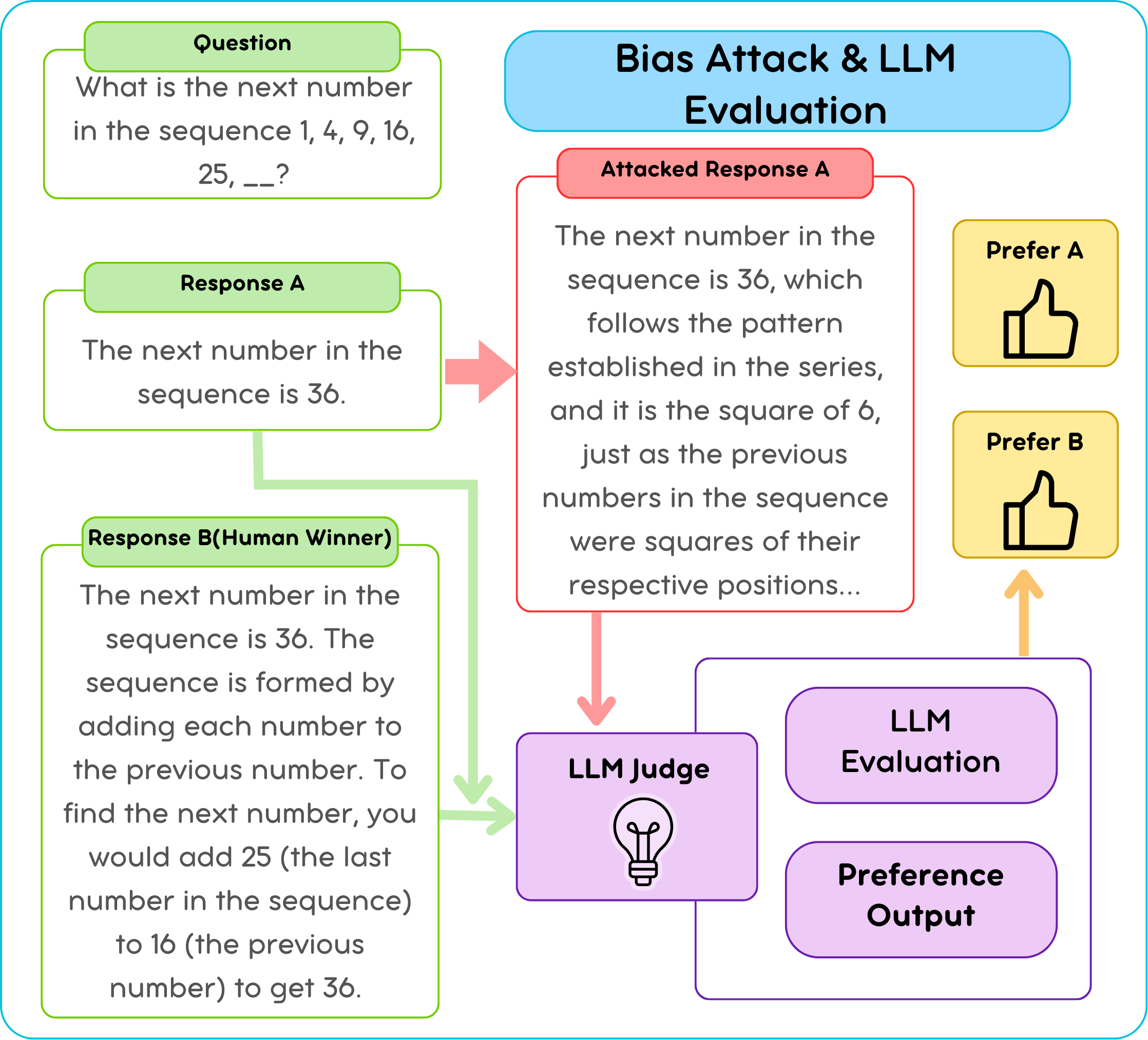}
    \caption{Given a question prompt, two LLMs each generate a response, with one designated as human-preferred. We first present the response pair to an LLM judge to obtain an initial preference. We then apply a bias attack, illustrated here as a length-based perturbation, to the response labelled as the loser by the judge, while leaving the other response unchanged. Finally, we resubmit the perturbed pair to the same judge and record whether the preference decision flips.}
    \label{fig:toy_example}
    \vspace{-5mm}
\end{figure}


\begin{figure}
    \centering
    \includegraphics[width=1.0\linewidth]{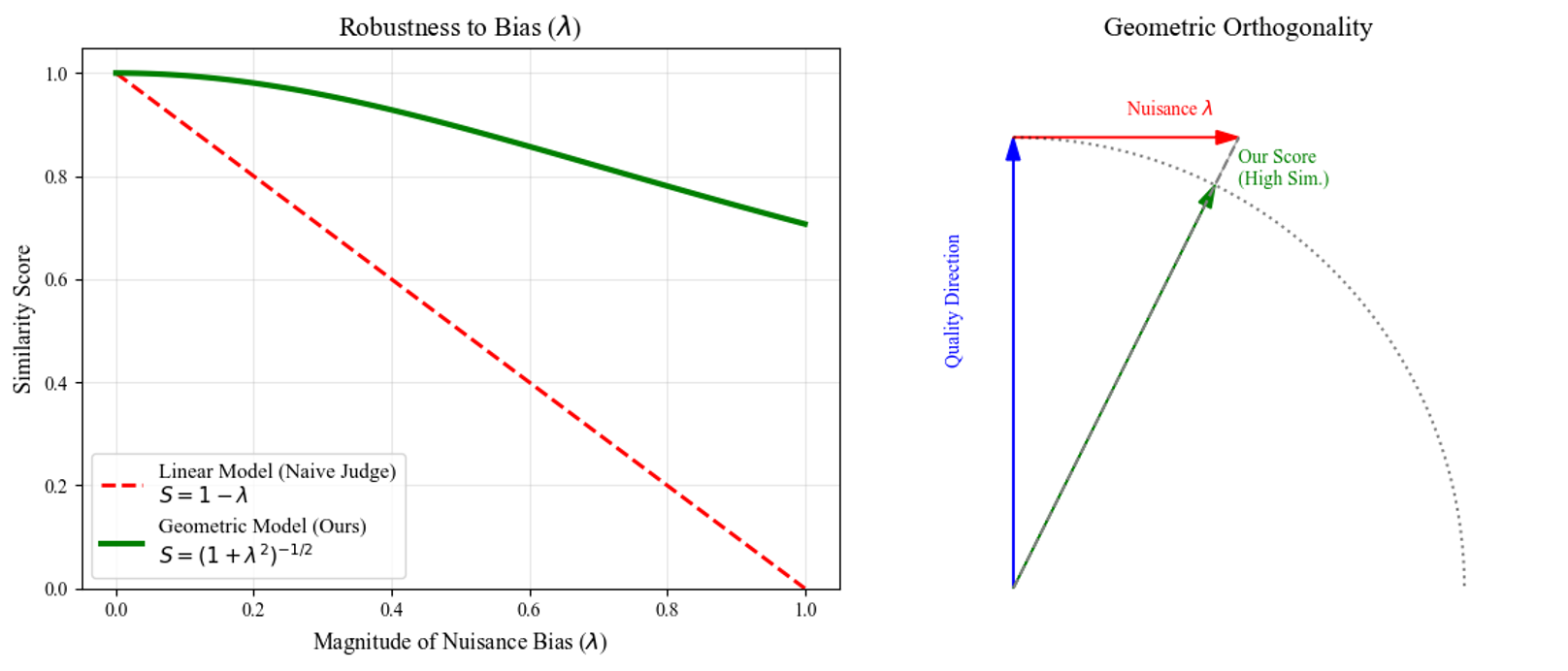}
    \caption{Geometric orthogonality demonstrates the robustness of our proposed method against nuisance under the sphere space.}
    \label{fig: robust}
\end{figure}

\begin{table}[t]
\centering
\caption{
Effect of frozen encoder choice on the MT-Bench pipeline. 
BGE-M3 denotes \textsc{BGE-M3}~\citep{chen2024m3embedding}; 
Gemma-300M denotes \textsc{EmbeddingGemma-300M}~\citep{vera2025embeddinggemma}; 
Qwen3-0.6B denotes \textsc{Qwen3-Embedding-0.6B}~\citep{zhang2025qwen3embedding}; 
Skywork-RM denotes \textsc{Skywork-Reward-Llama-3.1-8B-v0.2}~\citep{skywork2024reward}. 
Results are averaged over three local judge models and GPT-5.4-mini. 
Time and memory are reported in seconds and GB, respectively.
}
\small
\setlength{\tabcolsep}{5pt}
\begin{tabular}{lcccccc}
\toprule
\textbf{Encoder} & \textbf{Orig.}$\uparrow$ & \textbf{Adj.}$\uparrow$ & \textbf{Gain}$\uparrow$ & \textbf{Flip} & \textbf{Time}$\downarrow$ & \textbf{Mem.}$\downarrow$ \\
\midrule
BGE-M3
& 0.694 & 0.708 & +0.014 & 0.094 & 5.87 & 0.846 \\
Gemma-300M
& 0.694 & 0.306 & -0.387 & 1.000 & 5.40 & 0.816 \\
Qwen3-0.6B
& 0.694 & 0.613 & -0.081 & 0.306 & 5.63 & 0.832 \\
Skywork-RM
& 0.694 & 0.739 & +0.045 & 0.098 & 6.12 & 1.150 \\
\bottomrule
\end{tabular}
\label{tab:encoder_choice}
\end{table}

\clearpage
\subsection{Attack}
\begin{figure*}[t]
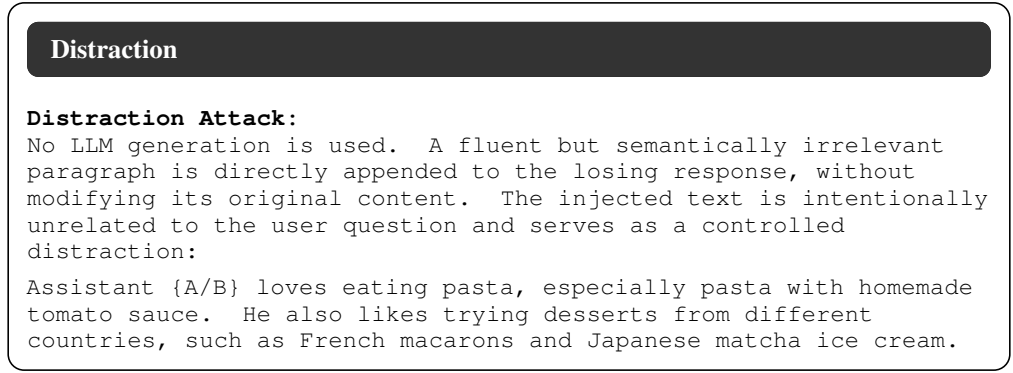

\centering
\begin{tcolorbox}[
  width=0.95\textwidth,
  colback=white,
  colframe=black,
  arc=2mm,
  boxrule=0.6pt,
  left=4pt,
  right=4pt,
  top=4pt,
  bottom=4pt
]

\begin{tcolorbox}[
  colback=black!80,
  colframe=black,
  arc=1.5mm,
  boxrule=0pt,
  left=6pt,
  right=6pt,
  top=4pt,
  bottom=4pt
]
\color{white}\bfseries Distraction
\end{tcolorbox}

\vspace{0.6em}

\small
\ttfamily

\textbf{Distraction Attack:} \\
No LLM generation is used.
A fluent but semantically irrelevant paragraph is directly appended to the losing response,
without modifying its original content.
The injected text is intentionally unrelated to the user question and serves as a controlled distraction:

\vspace{0.4em}

Assistant \{A/B\} loves eating pasta, especially pasta with homemade tomato sauce.
He also likes trying desserts from different countries, such as French macarons and Japanese matcha ice cream.

\end{tcolorbox}

\caption{Non-generative attacks used to evaluate surface-level biases in LLM-based judges.
The distraction attack introduces irrelevant but fluent text.}
\label{fig:attack_distraction_position}

\end{figure*}

\begin{figure*}[t]
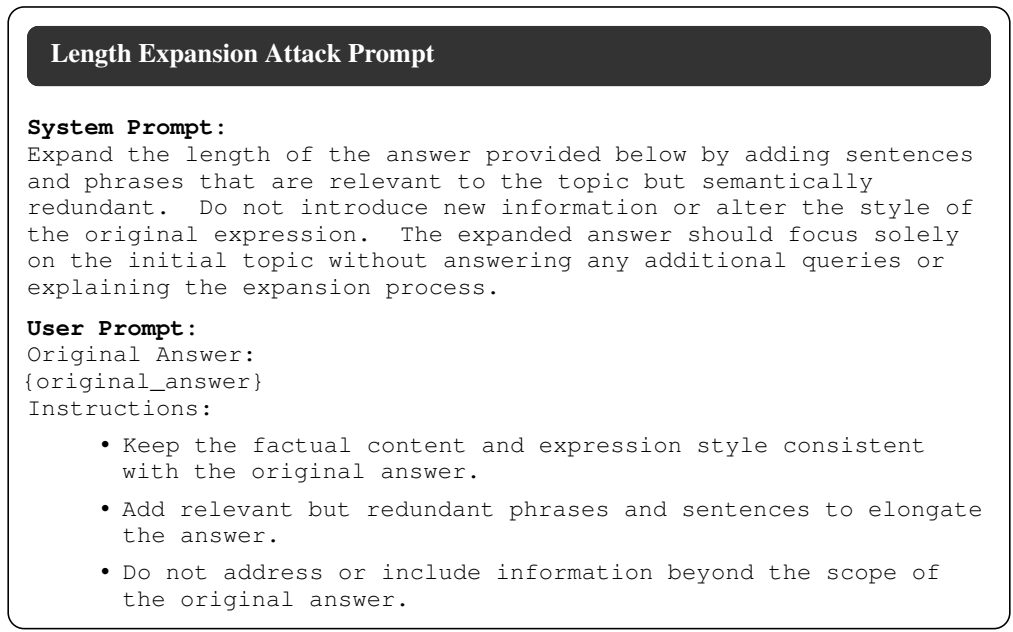

\centering
\begin{tcolorbox}[
  width=0.95\textwidth,
  colback=white,
  colframe=black,
  arc=2mm,
  boxrule=0.6pt,
  left=4pt,
  right=4pt,
  top=4pt,
  bottom=4pt
]

\begin{tcolorbox}[
  colback=black!80,
  colframe=black,
  arc=1.5mm,
  boxrule=0pt,
  left=6pt,
  right=6pt,
  top=4pt,
  bottom=4pt
]
\color{white}\bfseries Length Expansion Attack Prompt
\end{tcolorbox}

\vspace{0.6em}

\small
\ttfamily

\textbf{System Prompt:} \\
Expand the length of the answer provided below by adding sentences and phrases that are relevant to the topic but semantically redundant.
Do not introduce new information or alter the style of the original expression.
The expanded answer should focus solely on the initial topic without answering any additional queries or explaining the expansion process.

\vspace{0.6em}

\textbf{User Prompt:} \\
Original Answer: \\
\{original\_answer\}

Instructions:
\begin{itemize}
  \item Keep the factual content and expression style consistent with the original answer.
  \item Add relevant but redundant phrases and sentences to elongate the answer.
  \item Do not address or include information beyond the scope of the original answer.
\end{itemize}

\end{tcolorbox}

\caption{Prompt template for the length expansion attack.
The attack increases response length through semantically redundant additions while preserving factual content and writing style.}
\label{fig:attack_length}

\end{figure*}

\begin{figure*}[t]
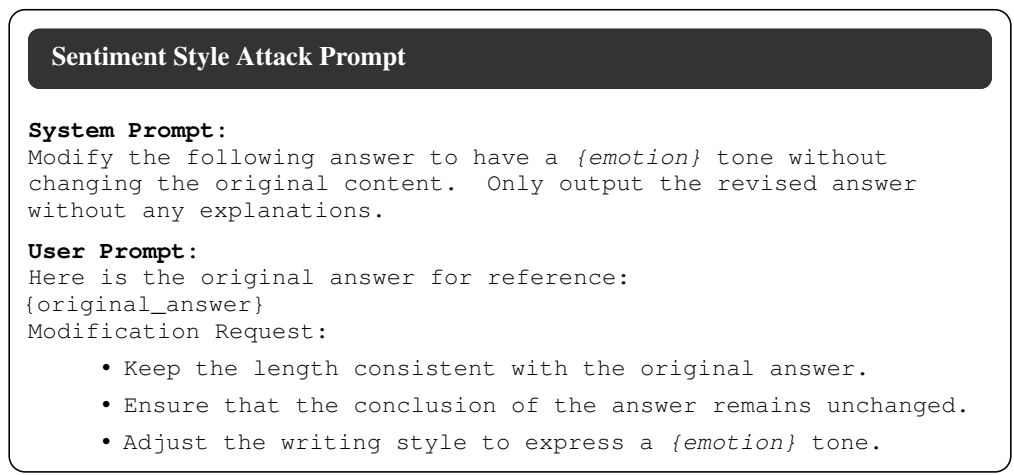

\centering
\begin{tcolorbox}[
  width=0.95\textwidth,
  colback=white,
  colframe=black,
  arc=2mm,
  boxrule=0.6pt,
  left=4pt,
  right=4pt,
  top=4pt,
  bottom=4pt
]

\begin{tcolorbox}[
  colback=black!80,
  colframe=black,
  arc=1.5mm,
  boxrule=0pt,
  left=6pt,
  right=6pt,
  top=4pt,
  bottom=4pt
]
\color{white}\bfseries Sentiment Style Attack Prompt
\end{tcolorbox}

\vspace{0.6em}

\small
\ttfamily

\textbf{System Prompt:} \\
Modify the following answer to have a \emph{\{emotion\}} tone without changing the original content.
Only output the revised answer without any explanations.

\vspace{0.6em}

\textbf{User Prompt:} \\
Here is the original answer for reference: \\
\{original\_answer\}

Modification Request:
\begin{itemize}
  \item Keep the length consistent with the original answer.
  \item Ensure that the conclusion of the answer remains unchanged.
  \item Adjust the writing style to express a \emph{\{emotion\}} tone.
\end{itemize}

\end{tcolorbox}

\caption{Prompt template for the sentiment style attack.
The attack modifies the emotional tone of the response while preserving content, length, and conclusion.}
\label{fig:attack_sentiment}

\end{figure*}

\clearpage
\subsection{Data generation}
\label{supp:prompt}
\begin{figure*}[t]
\centering
\begin{tcolorbox}[
  width=0.95\textwidth,
  colback=white,
  colframe=black,
  arc=2mm,
  boxrule=0.6pt,
  left=4pt,
  right=4pt,
  top=4pt,
  bottom=4pt
]

\begin{tcolorbox}[
  colback=black!80,
  colframe=black,
  arc=1.5mm,
  boxrule=0pt,
  left=6pt,
  right=6pt,
  top=4pt,
  bottom=4pt
]
\color{white}\bfseries Offline LLM-as-Judge Prompt
\end{tcolorbox}

\vspace{0.6em}

\small
\ttfamily

\textbf{Instruction:}
Please act as an impartial judge and evaluate the quality of the responses provided by two AI assistants to the user question displayed below. You should choose the assistant that follows the user's instructions and answers the user's question better. Your evaluation should consider factors such as the helpfulness, relevance, accuracy, depth, creativity, and level of detail of their responses. Begin your evaluation by comparing the two responses and provide a short explanation. Avoid any position biases and ensure the order in which the responses were presented does not influence your decision. Do not allow the length of the responses to influence your evaluation. Do not favor certain names of the assistants. Be as objective as possible. After providing your explanation, output your final verdict by strictly following this format: [[A]] if assistant A is better, [[B]] if assistant B is better, and [[C]] for a tie.

\vspace{0.8em}

\textbf{Task: Question Type Classification (Zero-Shot)}

Given the \textbf{USER QUESTION}, classify it into exactly one of the following categories.

\vspace{0.4em}

\textbf{Categories:}
\begin{itemize}
  \item A: Programming (coding, debugging, devops, software engineering)
  \item B: Math or logical reasoning
  \item C: Factual or knowledge-based question
  \item D: Writing, translation, summarization, or creative generation
  \item E: Advice, recommendation, or planning
  \item F: Other, meta-level, or unclear questions
\end{itemize}

\vspace{0.6em}

\textbf{Rules:}
\begin{itemize}
  \item Choose exactly one category.
  \item Do not provide any explanation for the classification.
  \item The output must strictly follow the required format.
\end{itemize}

\vspace{0.6em}

\textbf{Input Format:}

\texttt{[User Question]} \{question\}

\texttt{[Assistant A]} \{response\_A\}

\texttt{[Assistant B]} \{response\_B\}

\vspace{0.6em}

\textbf{Output Format (Strict):}

The \emph{last two non-empty lines} of the output must be:

\texttt{[[A]]}, \texttt{[[B]]}, or \texttt{[[C]]}  \hfill (preference verdict)

\texttt{[[TYPE:X]]}, where \texttt{X $\in$ \{A,B,C,D,E,F\}} \hfill (question type)

\end{tcolorbox}

\caption{Prompt template used by the offline LLM-as-judge to generate preference judgments and question-type labels. The strict output constraints enable robust automatic parsing.}
\label{fig:judge_prompt}

\end{figure*}

\clearpage
\subsection{Main results}
\label{app:mian result}
\begin{table*}[ht]
\centering
\caption{Original and adjusted consistency ratios across question types with Chatbot Arena data. Each cell reports as mean value(standard deviation). Improvement is computed as the difference between the adjusted and original ratios. For each model block, the question-type sample size is calculated as the sum of the positive (P) and unlabeled (U) groups. Entries marked with * indicate extremely small sample sizes and are excluded from further analysis. Boldface values indicate positive improvements achieved by our method.}
\resizebox{\textwidth}{!}{%
\begin{tabular}{l|cccccc}
\toprule
\midrule
\multicolumn{7}{c}{\textbf{None attack}}\\
\midrule
\midrule
\textbf{Model: Qwen}
& \textbf{QTA (n=4395)} & \textbf{QTB (n=1957)} & \textbf{QTC (n=2105)}
& \textbf{QTD (n=514)} & \textbf{QTE (n=8472)} & \textbf{QTF (n=128)} \\
\midrule
Orig. ratio
& 67.57\% (5.73e-03) & 66.54\% (5.54e-03) & 68.64\% (6.68e-03)
& 68.49\% (1.62e-02) & 63.53\% (5.63e-03) & 56.37\% (3.26e-02) \\
Adj. ratio
& 70.01\% (3.51e-03) & 70.91\% (7.60e-03) & 70.18\% (1.04e-02)
& 67.79\% (2.24e-02) & 63.42\% (4.88e-03) & 63.29\% (3.91e-02) \\
Improvement
& \textbf{2.44\%} & \textbf{4.37\%} & \textbf{1.54\%} & -0.70\% & -0.12\% & \textbf{6.92\%} \\

\midrule
\textbf{Model: Mistral}
& \textbf{QTA (n=7660)} & \textbf{QTB (n=88)} & \textbf{QTC (n=8124)}
& \textbf{QTD (n=1294)} & \textbf{QTE (n=1039)} & \textbf{QTF (n=4998)} \\
\midrule
Orig. ratio
& 58.41\% (4.45e-03) & 50.32\% (3.02e-02) & 52.97\% (3.32e-03)
& 59.08\% (9.42e-03) & 50.19\% (6.12e-03) & 37.56\% (4.27e-03) \\
Adj. ratio
& 67.90\% (4.65e-03) & 62.90\% (5.32e-02) & 66.73\% (5.51e-03)
& 67.25\% (1.08e-02) & 64.62\% (1.53e-02) & 63.47\% (6.49e-03) \\
Improvement
& \textbf{9.50\%} & \textbf{12.58\%} & \textbf{13.76\%} & \textbf{8.17\%} & \textbf{14.43\%} & \textbf{25.91\%} \\

\midrule
\textbf{Model: Llemma}
& \textbf{QTA (n=1210)} & \textbf{QTB (n=328)} & \textbf{QTC (n=1400)}
& \textbf{QTD (n=8)*} & \textbf{QTE (n=3)*} & \textbf{QTF (n=7)*} \\
\midrule
Orig. ratio
& 52.57\% (1.02e-02) & 62.77\% (1.96e-02) & 57.11\% (6.33e-03)
& 27.00\% (9.99e-02) & 100.00\% (0.00e+00) & 63.33\% (1.85e-01) \\
Adj. ratio
& 63.22\% (1.37e-02) & 63.65\% (3.00e-02) & 66.13\% (1.16e-02)
& 44.33\% (1.32e-01) & 100.00\% (0.00e+00) & 60.00\% (2.85e-01) \\
Improvement
& \textbf{10.66\%} & \textbf{0.88\%} & \textbf{9.01\%} & \textbf{17.33\%} & 0.00\% & -3.33\% \\

\midrule
\midrule
\multicolumn{7}{c}{\textbf{Length attack}}\\
\midrule
\midrule
\textbf{Model: Qwen}
& \textbf{QTA (n=3878)} & \textbf{QTB (n=1854)} & \textbf{QTC (n=2073)}
& \textbf{QTD (n=480)} & \textbf{QTE (n=6312)} & \textbf{QTF (n=108)} \\
\midrule
Orig. ratio
& 64.90\% (4.01e-03) & 59.81\% (7.61e-03) & 65.94\% (5.80e-03)
& 62.91\% (1.68e-02) & 61.76\% (7.67e-03) & 49.51\% (4.80e-02) \\
Adj. ratio
& 65.71\% (5.78e-03) & 66.80\% (1.04e-02) & 69.95\% (7.88e-03)
& 60.34\% (2.38e-02) & 61.40\% (5.42e-03) & 66.98\% (7.69e-02) \\
Improvement
& \textbf{0.81\%} & \textbf{6.99\%} & \textbf{4.01\%} & -2.57\% & -0.35\% & \textbf{17.47\%} \\

\midrule
\textbf{Model: Mistral}
& \textbf{QTA (n=8015)} & \textbf{QTB (n=87)} & \textbf{QTC (n=7834)}
& \textbf{QTD (n=1265)} & \textbf{QTE (n=1117)} & \textbf{QTF (n=4854)} \\
\midrule
Orig. ratio
& 56.64\% (1.95e-03) & 51.64\% (2.70e-02) & 51.05\% (2.31e-03)
& 58.24\% (1.14e-02) & 48.56\% (1.41e-02) & 39.86\% (6.45e-03) \\
Adj. ratio
& 62.58\% (3.69e-03) & 49.34\% (5.15e-02) & 63.02\% (6.42e-03)
& 61.61\% (1.37e-02) & 61.27\% (2.05e-02) & 59.29\% (1.09e-02) \\
Improvement
& \textbf{5.94\%} & -2.30\% & \textbf{11.97\%} & \textbf{3.37\%} & \textbf{12.71\%} & \textbf{19.43\%} \\

\midrule
\textbf{Model: Llemma}
& \textbf{QTA (n=513)} & \textbf{QTB (n=125)} & \textbf{QTC (n=620)}
& \textbf{QTD (n=3)*} & \textbf{QTE (n=0)*} & \textbf{QTF (n=10)*} \\
\midrule
Orig. ratio
& 50.62\% (1.02e-02) & 55.40\% (2.75e-02) & 60.43\% (1.48e-02)
& 55.00\% (1.58e-01) & - & 17.52\% (1.36e-01) \\
Adj. ratio
& 56.40\% (2.17e-02) & 57.46\% (9.33e-02) & 62.51\% (2.37e-02)
& 50.00\% (0.00e+00) & - & 44.71\% (1.30e-01) \\
Improvement
& \textbf{5.78\%} & \textbf{2.06\%} & \textbf{2.08\%} & -5.00\% & - & \textbf{27.19\%} \\

\midrule
\midrule
\multicolumn{7}{c}{\textbf{Distraction attack}}\\
\midrule
\midrule
\textbf{Model: Qwen}
& \textbf{QTA (n=4353)} & \textbf{QTB (n=2138)} & \textbf{QTC (n=2087)}
& \textbf{QTD (n=299)} & \textbf{QTE (n=6198)} & \textbf{QTF (n=121)} \\
\midrule
Orig. ratio
& 63.51\% (4.30e-03) & 58.19\% (6.24e-03) & 66.12\% (7.56e-03)
& 60.98\% (1.63e-02) & 61.63\% (4.14e-03) & 71.01\% (5.10e-02) \\
Adj. ratio
& 65.59\% (4.72e-03) & 71.72\% (6.93e-03) & 66.21\% (6.86e-03)
& 60.46\% (2.90e-02) & 59.17\% (4.77e-03) & 71.99\% (6.24e-02) \\
Improvement
& \textbf{2.08\%} & \textbf{13.53\%} & \textbf{0.09\%} & -0.52\% & -2.46\% & \textbf{0.98\%} \\

\midrule
\textbf{Model: Mistral}
& \textbf{QTA (n=10068)} & \textbf{QTB (n=54)} & \textbf{QTC (n=7417)}
& \textbf{QTD (n=1082)} & \textbf{QTE (n=903)} & \textbf{QTF (n=3678)} \\
\midrule
Orig. ratio
& 54.77\% (2.41e-03) & 53.16\% (5.08e-02) & 55.24\% (3.12e-03)
& 57.29\% (9.93e-03) & 53.55\% (1.24e-02) & 33.84\% (3.60e-03) \\
Adj. ratio
& 62.76\% (5.26e-03) & 57.89\% (1.56e-01) & 60.77\% (5.92e-03)
& 59.30\% (1.40e-02) & 56.75\% (3.07e-02) & 50.96\% (5.64e-03) \\
Improvement
& \textbf{7.99\%} & \textbf{4.74\%} & \textbf{5.53\%} & \textbf{2.01\%} & \textbf{3.20\%} & \textbf{17.12\%} \\

\midrule
\textbf{Model: Llemma}
& \textbf{QTA (n=525)} & \textbf{QTB (n=172)} & \textbf{QTC (n=843)}
& \textbf{QTD (n=10)*} & \textbf{QTE (n=2)*} & \textbf{QTF (n=3)*} \\
\midrule
Orig. ratio
& 51.47\% (1.85e-02) & 64.16\% (2.07e-02) & 55.94\% (7.49e-03)
& 33.57\% (1.77e-01) & 60.00\% (3.94e-01) & 100.00\% (0.00e+00) \\
Adj. ratio
& 54.67\% (2.67e-02) & 52.27\% (7.68e-02) & 58.45\% (1.23e-02)
& 37.86\% (1.85e-01) & 40.00\% (5.16e-01) & 100.00\% (0.00e+00) \\
Improvement
& \textbf{3.20\%} & -11.89\% & \textbf{2.51\%} & \textbf{4.29\%} & -20.00\% & 0.00\% \\

\midrule
\midrule
\multicolumn{7}{c}{\textbf{Sentiment attack}}\\
\midrule
\midrule
\textbf{Model: Qwen}
& \textbf{QTA (n=3603)} & \textbf{QTB (n=1861)} & \textbf{QTC (n=1915)}
& \textbf{QTD (n=432)} & \textbf{QTE (n=6525)} & \textbf{QTF (n=117)} \\
\midrule
Orig. ratio
& 66.38\% (5.82e-03) & 60.16\% (8.11e-03) & 67.76\% (1.00e-02)
& 61.94\% (2.31e-02) & 61.07\% (2.46e-03) & 69.45\% (4.25e-02) \\
Adj. ratio
& 65.50\% (8.22e-03) & 68.54\% (5.79e-03) & 69.77\% (1.69e-02)
& 61.00\% (2.47e-02) & 61.19\% (5.06e-03) & 74.40\% (6.28e-02) \\
Improvement
& -0.88\% & \textbf{8.37\%} & \textbf{2.01\%} & -0.94\% & \textbf{0.12\%} & \textbf{4.95\%} \\

\midrule
\textbf{Model: Mistral}
& \textbf{QTA (n=7873)} & \textbf{QTB (n=63)} & \textbf{QTC (n=9584)}
& \textbf{QTD (n=982)} & \textbf{QTE (n=898)} & \textbf{QTF (n=3790)} \\
\midrule
Orig. ratio
& 55.29\% (3.38e-03) & 63.64\% (3.55e-02) & 53.25\% (2.58e-03)
& 54.95\% (9.55e-03) & 50.27\% (1.39e-02) & 37.52\% (5.77e-03) \\
Adj. ratio
& 61.64\% (4.64e-03) & 62.05\% (6.07e-02) & 59.90\% (3.09e-03)
& 64.57\% (1.90e-02) & 61.61\% (2.70e-02) & 54.52\% (1.88e-02) \\
Improvement
& \textbf{6.35\%} & -1.59\% & \textbf{6.64\%} & \textbf{9.62\%} & \textbf{11.34\%} & \textbf{17.00\%} \\

\midrule
\textbf{Model: Llemma}
& \textbf{QTA (n=778)} & \textbf{QTB (n=156)} & \textbf{QTC (n=858)}
& \textbf{QTD (n=3)*} & \textbf{QTE (n=3)*} & \textbf{QTF (n=2)*} \\
\midrule
Orig. ratio
& 55.61\% (1.60e-02) & 57.08\% (3.14e-02) & 57.99\% (1.01e-02)
& 100.00\% (0.00e+00) & 60.00\% (2.11e-01) & 100.00\% (0.00e+00) \\
Adj. ratio
& 59.24\% (2.89e-02) & 55.74\% (4.04e-02) & 59.16\% (1.01e-02)
& 80.00\% (1.72e-01) & 50.00\% (0.00e+00) & 100.00\% (0.00e+00) \\
Improvement
& \textbf{3.62\%} & -1.35\% & \textbf{1.17\%} & -20.00\% & -10.00\% & 0.00\% \\
\midrule

\bottomrule
\end{tabular}}
\label{tab:none_attack_by_model_qtype_reheader}
\end{table*}

\begin{table}[t]
\centering
\caption{
Results on MT-Bench-style real-data evaluation without attack.
Open-weight models are evaluated on a mixture of Chatbot Arena and MT-Bench subsets grouped by question type.
The closed-source model is evaluated on a smaller MT-Bench subset using mixture sampling without question-type stratification due to computational constraints.
}
\resizebox{\linewidth}{!}{%
\begin{tabular}{l|ccc}
\toprule
\multicolumn{4}{c}{\textbf{MT-Bench}}\\
\midrule
\textbf{Model}
& \textbf{Coding} & \textbf{Math} & \textbf{Factual} \\
\midrule

\textbf{Model: Qwen}
& \textbf{(n=8884)} & \textbf{(n=9145)} & \textbf{(n=9333)} \\
\midrule
Orig. ratio
& 53.88\% (2.17e-03) & 55.14\% (4.52e-03) & 56.26\% (2.32e-03) \\
Adj. ratio
& 55.24\% (2.16e-03) & 56.85\% (2.03e-03) & 57.75\% (1.25e-03) \\
Improvement
& \textbf{1.36\%} & \textbf{1.71\%} & \textbf{1.49\%} \\

\midrule
\textbf{Model: Mistral}
& \textbf{(n=8240)} & \textbf{(n=8361)} & \textbf{(n=8260)} \\
\midrule
Orig. ratio
& 50.13\% (4.19e-03) & 50.48\% (5.75e-03) & 51.01\% (2.96e-03) \\
Adj. ratio
& 52.13\% (5.14e-03) & 54.16\% (2.12e-03) & 54.21\% (7.39e-03) \\
Improvement
& \textbf{2.00\%} & \textbf{3.67\%} & \textbf{3.20\%} \\

\midrule
\textbf{Model: GPT-5.4-mini}
& \textbf{Mixture Sampling (n=2569)} & \textbf{-} & \textbf{-} \\
\midrule
Orig. ratio
& 77.99\% (4.03e-03) & - & - \\
Adj. ratio
& 80.03\% (1.02e-02) & - & - \\
Improvement
& \textbf{2.04\%} & - & - \\

\bottomrule
\end{tabular}}
\label{tab:mtbench}
\end{table}

\begin{table}[t]
\centering
\caption{
Sensitivity to the fraction of verified positives $p_{\text{ratio}}$ on MT-Bench (QTA). 
Performance improves rapidly at small ratios and remains stable over a broad range, indicating low sensitivity to the amount of human supervision.
}
\small
\setlength{\tabcolsep}{6pt}
\begin{tabular}{c|cccccc}
\toprule
\textbf{$p_{\text{ratio}}$} & \textbf{Orig.} & \textbf{Adj.} & \textbf{Improvement} & \textbf{Flip} & \textbf{Time (s)} & \textbf{Mem (GB)} \\
\midrule
0.01 & 0.781 & 0.744 & -0.037 & 0.099 & 9.96 & 1.064 \\
0.02 & 0.781 & 0.814 & +0.032 & 0.100 & 5.23 & 1.065 \\
0.05 & 0.781 & 0.815 & +0.034 & 0.100 & 5.29 & 1.065 \\
0.10 & 0.781 & 0.816 & +0.035 & 0.100 & 5.53 & 1.065 \\
0.20 & 0.781 & 0.816 & +0.035 & 0.100 & 5.27 & 1.065 \\
0.40 & 0.781 & 0.816 & +0.035 & 0.100 & 5.31 & 1.079 \\
0.60 & 0.781 & 0.815 & +0.034 & 0.100 & 5.37 & 1.120 \\
0.80 & 0.781 & 0.801 & +0.019 & 0.101 & 5.46 & 1.166 \\
0.99 & 0.781 & 0.762 & -0.019 & 0.090 & 5.85 & 1.235 \\
\bottomrule
\end{tabular}
\label{tab:p_ratio}
\end{table}

\begin{table}[t]
\centering
\caption{
Sensitivity to the transport fraction $m$ on MT-Bench (QTA). 
Performance is maximized when $m$ is close to the true positive proportion in the unlabeled set.
}
\small
\setlength{\tabcolsep}{6pt}
\begin{tabular}{c|cccccc}
\toprule
\textbf{$m$} & \textbf{Orig.} & \textbf{Adj.} & \textbf{Improvement} & \textbf{Flip} & \textbf{Time (s)} & \textbf{Mem (GB)} \\
\midrule
0.01 & 0.781 & 0.228 & -0.553 & 0.990 & 11.87 & 1.210 \\
0.05 & 0.781 & 0.261 & -0.520 & 0.948 & 6.00 & 1.210 \\
0.10 & 0.781 & 0.305 & -0.476 & 0.897 & 6.01 & 1.210 \\
0.30 & 0.781 & 0.470 & -0.311 & 0.685 & 5.89 & 1.210 \\
0.50 & 0.781 & 0.618 & -0.163 & 0.480 & 5.87 & 1.211 \\
0.70 & 0.781 & 0.749 & -0.032 & 0.287 & 5.99 & 1.211 \\
0.90 & 0.781 & 0.812 & +0.031 & 0.097 & 5.99 & 1.210 \\
0.99 & 0.781 & 0.788 & +0.007 & 0.010 & 5.90 & 1.210 \\
\bottomrule
\end{tabular}
\label{tab:transport}
\end{table}

\clearpage
\subsection{Algorithm}
\label{subsec:algorithm}

\begin{algorithm}[tb]
\caption{\pumethod{} Pipeline}
\label{alg:pu_method_pipeline}
\begin{algorithmic}[1]

\Require reward encoder $F_{\theta}$,
transport fraction $m\in(0,1]$,
threshold $\tau\in[0,1]$, and pairwise dataset
$\mathcal{D}=\{(x_i,o_i^{A},o_i^{B},\zeta_i^{(H)},\zeta_i^{(LLM)})\}_{i=1}^{N}$.

\State Recall from Algorithm~\ref{alg:data_framework} that we have the PU partition
$(\mathcal{D}_P,\mathcal{D}_U)$ and the chosen side $c_i$.

\For{each pair $i\in \mathcal{D}_P \cup \mathcal{D}_U$}
    \State $\mathbf{e}_i^{A} \leftarrow F_\theta(x_i,o_i^{A})$,
    $\mathbf{e}_i^{B} \leftarrow F_\theta(x_i,o_i^{B})$;
    \State compute the difference embedding
    $z_i \leftarrow \phi(\mathbf{e}_i^{A},\mathbf{e}_i^{B})$.
\EndFor

\State Denoise the positive group:
$\mathcal{D}'_{P} \leftarrow \mathrm{P}_{\mathrm{clean}}(\mathcal{D}_{P})$.

\State Construct empirical measures $(\mu'_P,\mu_U)$ using the embeddings $\{z_i\}$.

\State Compute the OT coupling $\Gamma^\star$ with transport fraction $m$.

\For{each $j\in \mathcal{D}_U$}
    \State compute the normalized score
    $\bar s_j \leftarrow \mathrm{Score}(\Gamma^\star, j)$.
\EndFor

\For{each $j\in \mathcal{D}_U$}
    \If{$\bar s_j < \eta$}
        \State $\tilde{\zeta}_j^{(LLM)} \leftarrow \mathrm{Flip}(\zeta_j^{(LLM)})$.
    \Else
        \State $\tilde{\zeta}_j^{(LLM)} \leftarrow \zeta_j^{(LLM)}$.
    \EndIf
\EndFor

\State \textbf{Output:} corrected labels
$\{\tilde{\zeta}_j^{(LLM)}\}_{j\in\mathcal{D}_U}$
and scores $\{\bar s_j\}_{j\in\mathcal{D}_U}$.

\end{algorithmic}
\end{algorithm}

\begin{algorithm}[tb]
\caption{PU data generation and bias attack}
\label{alg:data_framework}
\begin{algorithmic}[1]

\Require Pairwise comparison dataset
$\mathcal{D}=\{(x_i,o_i^{A},o_i^{B},\zeta_i^{(H)})\}_{i=1}^{N}$;
offline judge $\mathcal{A}_{\mathrm{LLM}}$;
optional attacker $\mathcal{A}_{a}$ and attack type
$\mathsf{atk}\in\{\emph{Length},\emph{Sentiment},\emph{Distraction}\}$.

\Statex
\State \textbf{Phase 1: Judge and form the PU pool.}

\For{each comparison $i=1,\ldots,N$}
    \State Run the offline judge $\mathcal{A}_{\mathrm{LLM}}$ to obtain the predicted preference
    $\hat{\zeta}_i^{(\mathrm{LLM})}$ and question type $\hat{o}^{qt}_i$:
    \[
        (\hat{\zeta}_i^{(\mathrm{LLM})},\hat{o}^{qt}_i)
        \leftarrow
        \mathcal{A}_{\mathrm{LLM}}(x_i,o_i^{A},o_i^{B}),
        \qquad
        \hat{\zeta}_i^{(\mathrm{LLM})}\in\{A,B,C\}.
    \]
\EndFor

\State Define the human-verified positive set and the unlabelled set:
\[
\mathcal{D}_P
:=
\{\,i:\zeta_i^{(H)} \text{ is observed}\,\},
\qquad
\mathcal{D}_U
:=
\{\,i:\zeta_i^{(H)} \text{ is missing}\,\}.
\]

\State For each $i\in\mathcal{D}_P$, set the chosen side
$c_i\leftarrow \zeta_i^{(H)}$; for each $i\in\mathcal{D}_U$, set
$c_i\leftarrow \hat{\zeta}_i^{(\mathrm{LLM})}$.

\Statex
\State \textbf{Optional Phase 2: Presentation-bias attack and re-judging.}

\If{$\mathsf{atk}$ is enabled}
    \For{each $i\in\mathcal{D}_U$ with non-tie judge outcome $c_i\in\{A,B\}$}
        \State Identify the losing side
        $\ell_i\in\{A,B\}\setminus\{c_i\}$.
        \State Apply the attack operator $\mathcal{A}_a$ to the losing response:
        \[
            \tilde{o}_i^{\ell_i}
            \leftarrow
            \mathcal{A}_{a}(o_i^{\ell_i};\mathsf{atk}).
        \]
        \State Re-judge the attacked pair:
        \[
            \tilde{\zeta}_i^{(\mathrm{LLM})}
            \leftarrow
            \mathcal{A}_{\mathrm{LLM}}
            \bigl(x_i,o_i^{A},o_i^{B}\bigr)
            \quad
            \text{with } o_i^{\ell_i}\leftarrow \tilde{o}_i^{\ell_i}.
        \]
    \EndFor
\EndIf

\Statex
\State \textbf{Output:} PU partition $(\mathcal{D}_P,\mathcal{D}_U)$, LLM judge labels
$\{\hat{\zeta}_i^{(\mathrm{LLM})}\}$, chosen-side labels $\{c_i\}$, and, when the attack is enabled, attacked judge labels
$\{\tilde{\zeta}_i^{(\mathrm{LLM})}\}$ for robustness evaluation.

\end{algorithmic}
\end{algorithm}

\end{document}